\documentclass{article}

\pdfoutput=1
    \PassOptionsToPackage{numbers, compress}{natbib}

\usepackage[preprint]{neurips_2026}
\makeatletter
\renewcommand{\@noticestring}{}
\makeatother

\usepackage[utf8]{inputenc} 
\usepackage[T1]{fontenc}    
\usepackage{hyperref}       
\usepackage{url}            
\usepackage{booktabs}       
\usepackage{amsfonts}       
\usepackage{nicefrac}       
\usepackage{microtype}      
\usepackage{xcolor}         

\usepackage{graphicx}
\usepackage{booktabs}

\usepackage[accsupp]{axessibility}  

\usepackage{hyperref}

\usepackage{orcidlink}

\usepackage{booktabs}
\usepackage{graphicx}
\usepackage[table]{xcolor}
\usepackage{pifont}
\usepackage{xspace}
\usepackage{colortbl}
\usepackage{multirow}
\usepackage{caption}
\usepackage{pifont}

\definecolor{GreenColor}{rgb}{0.137,0.573,0.565}
\definecolor{yzybest}{rgb}{0.98, 0.8, 0.8}
\definecolor{yzysecond}{rgb}{0.99, 0.88, 0.77}
\definecolor{tocred}{RGB}{149,0,0}

\newcommand{\firstc}{\cellcolor{yzybest}}
\newcommand{\secondc}{\cellcolor{yzysecond}}
\newcommand{\cg}{\color{gray!60}}

\title{\Large LIST3R: Long-sequence Instance-aware 3D Reconstruction}

\author{
Jing Gao,\negmedspace\textsuperscript{1}
Wei Wang,\negmedspace\textsuperscript{1, $\dagger$}
Feiran Wang,\negmedspace\textsuperscript{2}
Yan Yan\textsuperscript{2} \vspace{1ex} \\
\textsuperscript{1}Beijing Jiaotong University \qquad
\textsuperscript{2}University of Illinois Chicago
}

\begin{document}
\maketitle

\begingroup
\renewcommand{\thefootnote}{\fnsymbol{footnote}}
\footnotetext[2]{Corresponding Author.}
\endgroup

\vspace{-14pt}
\begin{figure*}[h]
  \centering
  \includegraphics[width=1\textwidth]{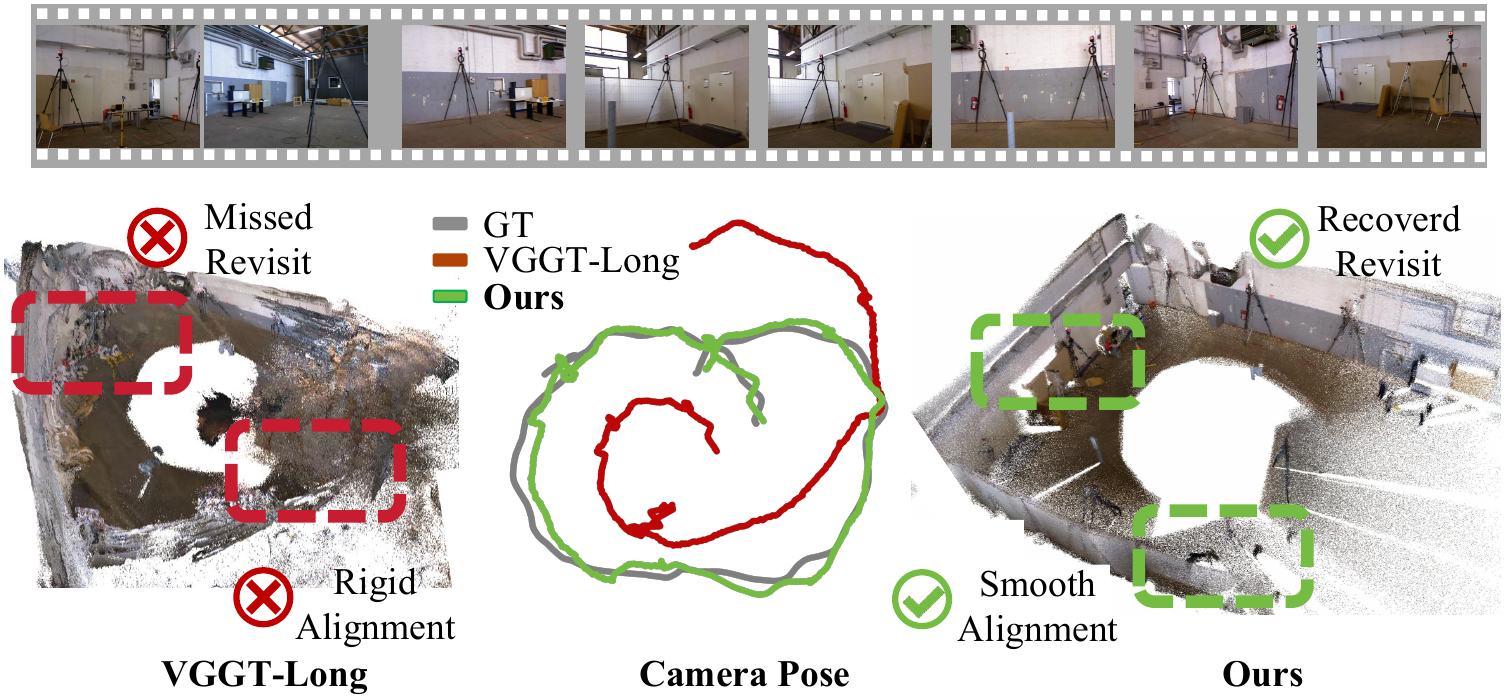}
  \caption{
  LIST3R leverages instance guidance to recover more effective revisits and smoother cross-subsequence alignment, producing more accurate and stable camera trajectories than the baseline.
}
  \label{fig:teaser}
\end{figure*}

\begin{abstract}
We present LIST3R, an instance-aware framework for long-sequence 3D reconstruction inspired by the way humans organize spatial memory around stable and recognizable objects.
LIST3R organizes long-sequence reconstruction around instance anchors, using them to reconnect fragmented subsequences and consolidate local observations into a coherent global 3D scene.
Given a long video,  our approach partitions it into overlapping subsequences and builds a structured local instance library for each partial reconstruction, maintaining persistent trackable anchors with semantic and geometric evidence.
These anchors are matched across subsequences to recover revisited regions and provide object-aware constraints for fragment alignment, producing a consistent global reconstruction.
During this process, the evolving geometric evidence updates the local instance libraries and progressively organizes them into a unified global 3D instance library.
Experiments on long-sequence benchmarks show that our method produces more accurate trajectories and higher-quality 3D reconstructions, highlighting the effectiveness of persistent instance anchors for organizing long-horizon 3D reconstruction.
Our code is available on the project page: \url{https://yixn965.github.io/LIST3R/}.
\end{abstract}

\section{Introduction}
\label{sec:intro}

Humans organize long visual experiences around a small number of stable and recognizable objects.
When navigating through an environment, these persistent cues help structure spatial memory, infer location, and reconnect observations made at different times~\cite{o1979precis,janzen_selective_2004,brunec2019cognitive,wang2026cognimap3d}.
Even under viewpoint changes, occlusion, or partial visibility, they remain effective for recognizing revisited places and maintaining a coherent understanding of the surrounding space~\cite{epstein_cortical_1998,dicarlo_untangling_2007,dicarlo_how_2012}.
Such an ability is particularly important for long-horizon visual understanding, where the challenge is not only to interpret the current observation, but also to consistently relate it to observations from the distant past.

Building on these insights, long-sequence 3D reconstruction calls for a framework that unifies three key capabilities:
1) maintaining persistent anchors that remain usable across time, viewpoint changes, and fragmented observations;
2) re-identifying revisited places over long temporal horizons; and
3) integrating partial reconstructions into a globally consistent representation.
These capabilities are essential for extending reconstruction beyond strong local geometry, enabling the system to preserve reliable long-range associations and consistent scene structure throughout the full sequence.

Recent work on long-sequence 3D reconstruction mainly follows two directions.
One line of work adopts streaming formulations~\cite{wang2025continuous,chen2025ttt3r}, where frames are processed sequentially while a compact latent memory is updated over time.
While such methods enable long-horizon processing, earlier scene information may be gradually forgotten as the sequence grows, making it difficult to preserve reliable long-range associations over extended videos.
The other line partitions a long video into overlapping subsequences, reconstructs each subsequence independently with a strong feed-forward foundation model, and then merges the partial results into a unified scene~\cite{deng2025vggt,xie2026scal3r}.
By retaining the full reconstruction power of the base model~\cite{wang2025vggt,wang2025pi,keetha2025mapanything} within each local window, this paradigm often yields high-fidelity local reconstructions for long videos.
However, these approaches still fall short of the capabilities needed to organize long visual experiences for consistent 3D reconstruction.

In this work, we propose \textbf{LIST3R}, an instance-aware framework for long-sequence 3D reconstruction.
We use object instances as persistent scene anchors to organize reconstruction across long temporal spans.
With these anchors, our method provides the reconstruction process with three key capabilities.
First, it establishes clear, reliable, and trackable local scene cues from fragmented observations.
Second, these cues allow the system to recall revisited regions across fragments and recover their spatial associations.
Third, the system continuously updates and integrates semantic and geometric evidence into a coherent reconstruction and a structured scene-level representation.
In this way, our approach complements existing long-sequence methods with stable organizing cues, enabling the system to organize long visual experiences for consistent 3D reconstruction.

Specifically, given a long video, we first partition it into overlapping subsequences and reconstructs each subsequence with a feed-forward 3D foundation model.
For each partial reconstruction, we initialize a local instance library by collecting trackable instance observations and associating them with reconstructed geometric evidence.
Using these local libraries, we first identify reliable long-range revisits from instance-level cues and selects the corresponding revisited subsequences.
It then jointly merges revisited and neighboring subsequences with instance-guided constraints, which are further refined by confidence-weighted optimization for global consistency.
As reconstruction proceeds, our method enriches instance evidence with updated geometry and consolidates local instances into a global 3D instance library, yielding a consistent scene-level representation.

Experiments on multiple long-sequence benchmarks show that our method improves camera pose accuracy and point-cloud reconstruction quality.
Compared with existing streaming and subsequence-based methods, our method achieves better global reconstruction consistency, demonstrating the effectiveness of persistent instance anchors for organizing long-horizon 3D reconstruction.

\section{Related Work}
\label{sec:related}

\subsection{Foundation Models for 3D Reconstruction}

Recent 3D reconstruction has been substantially advanced by feed-forward geometric foundation models that directly infer scene geometry and camera parameters from RGB images.
DUSt3R~\cite{wang2024dust3r} pioneered this direction by predicting aligned pointmaps from image pairs without requiring known poses or explicit geometric optimization.
MASt3R~\cite{leroy2024grounding} further strengthens this formulation with improved local matching, leading to more reliable correspondences and stronger geometric reconstruction.
Subsequent works extend this paradigm beyond pairwise reconstruction.
For example, Fast3R~\cite{yang2025fast3r} processes many views jointly in a single forward pass, reducing the need for costly pairwise matching and global alignment.
Recent multi-view foundation models such as VGGT~\cite{wang2025vggt}, $\pi^3$~\cite{wang2025pi}, and MapAnything~\cite{keetha2025mapanything} further regress camera parameters and dense 3D geometry from flexible multi-view inputs, substantially improving local multi-view reconstruction quality.
These models provide strong geometric priors for scene reconstruction, but their direct multi-view interaction remains difficult to scale to long video sequences due to memory and context-length limitations.

\subsection{Long-Sequence Reconstruction}

Extending feed-forward reconstruction to long sequences has emerged as an active research direction. Existing approaches can be categorized into two paradigms.
One direction adopts stateful or streaming formulations, where frames are processed sequentially while an evolving latent state is maintained over time.
CUT3R~\cite{wang2025continuous} represents this line with persistent state-based continuous 3D perception, while TTT3R~\cite{chen2025ttt3r} improves length generalization through test-time adaptation.
Although such methods avoid full-sequence global reasoning, earlier scene information may still be gradually weakened as the sequence grows, making robust long-range association difficult.

The other direction partitions a long sequence into local submaps, reconstructs them with a strong feed-forward model, and then aligns them globally.
VGGT-Long~\cite{deng2025vggt} scales VGGT~\cite{wang2025vggt} to kilometer-scale monocular RGB sequences through chunk-based processing, overlapping alignment, and lightweight loop-closure optimization.
Scal3R~\cite{xie2026scal3r} improves large-scale reconstruction with scalable test-time training and a neural global context representation for better long-range reasoning.
However, they still lack stable anchors that persist across subsequences, making long-range retrieval and cross-submap association unreliable under large viewpoint changes.

\section{Methods}
\label{sec:methods}

\begin{figure*}[t]
  \centering
  \includegraphics[width=\textwidth]{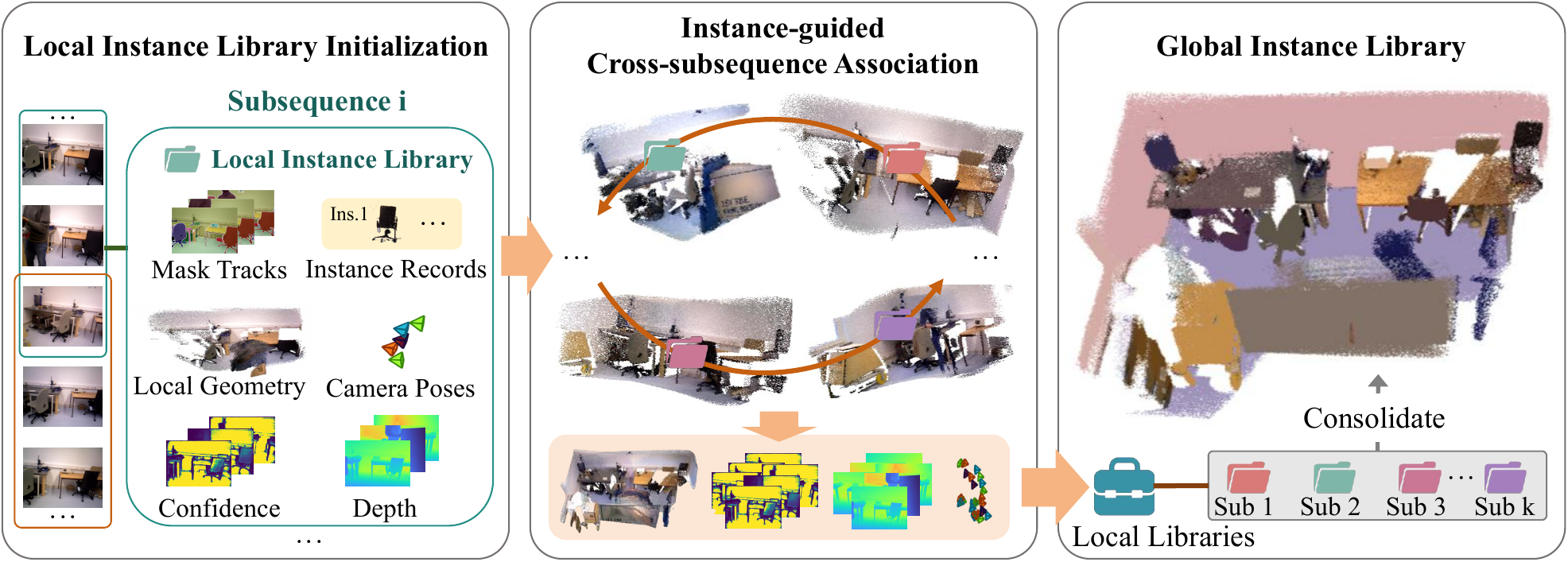}
  \vspace{-1.5em}
  \caption{\textbf{Method overview.}
  LIST3R first builds a local instance library for each subsequence to capture recognizable object cues.
  These cues are then used to connect subsequences and assemble fragmented local reconstructions into a coherent global scene.
  As reconstruction progresses, instance evidence is continuously updated and finally consolidated into a global instance library.
}
  \label{fig:method}
  \vspace{-1.0em}
\end{figure*}

Our core idea is to use recognizable object instances as persistent anchors throughout the reconstruction process.
As shown in Fig.~\ref{fig:method}, our method follows a three-stage pipeline.
First, we build a local instance library for each subsequence, organizing partial reconstructions around trackable object-centric cues (Sec.~\ref{sec:local_instance_initialization}).
Second, these local instance libraries are used to establish cross-subsequence associations, including long-range revisit discovery and instance-aware subsequence merging, which are further refined by confidence-weighted optimization for global consistency (Sec.~\ref{sec:instance_guided_association}).
Finally, local instance observations are consolidated into a unified 3D instance library (Sec.~\ref{sec:global_instance_library}).

\subsection{Local Instance Library Initialization}
\label{sec:local_instance_initialization}

Given a long image sequence $\mathcal{I}=\{I_t\}_{t=1}^{N}$, we first partition it into a set of overlapping local subsequences.
The $k$-th subsequence $\mathcal{S}^{(k)}$ contains $L$ frames, with $O$ frames overlapping adjacent subsequences.
To enable each local subsequence to provide object-level cues for later cross-subsequence association, we construct a local instance library within each subsequence.

To cover objects observed under different viewpoints, we uniformly select a set of anchor frames from each subsequence $\mathcal{S}^{(k)}$.
On these anchor frames, we perform lightweight query-based instance discovery, following the design philosophy of~\cite{ye2025entitysam}.
This process initializes each discovered object instance $q \in \{1,\ldots,Q_k\}$ with a mask prompt $\tilde{\mathbf{m}}_{q}^{(k)}$ and an object feature $\tilde{\mathbf{z}}_{q}^{(k)}$.

Using the initialized mask prompts $\{\tilde{\mathbf{m}}_{q}^{(k)}\}_{q=1}^{Q_k}$, we employ SAM3~\cite{carion2025sam} to propagate the discovered objects over the subsequence.
This produces a temporally consistent mask track for each instance.
Meanwhile, the corresponding object features and frame-level locations are aggregated to form a stable instance record.
The resulting local instance library of the $k$-th subsequence is defined as
\begin{equation}
\mathcal{L}_{\mathrm{loc}}^{(k)}
=
\left\{
\ell_i^{(k)}
\right\}_{i=1}^{N_k},
\qquad
\ell_i^{(k)}
=
\left(
a_i,
M_i,
V_i,
P_i,
\mathbf{z}_i
\right),
\end{equation}
where all attributes are defined within the $k$-th subsequence.
Specifically, $a_i$ is the local instance identifier,
$M_i$ denotes the multi-frame mask track,
$V_i$ is the set of visible frames,
$P_i$ stores the corresponding frame-level 2D locations, and
$\mathbf{z}_i$ denotes the aggregated local instance feature.
Through this process, each subsequence is organized into a local instance library, which stores trackable object-level anchors together with their associated attributes.

\subsection{Instance-guided Cross-subsequence Association}
\label{sec:instance_guided_association}

\begin{figure*}[t]
  \centering
  \includegraphics[width=\textwidth]{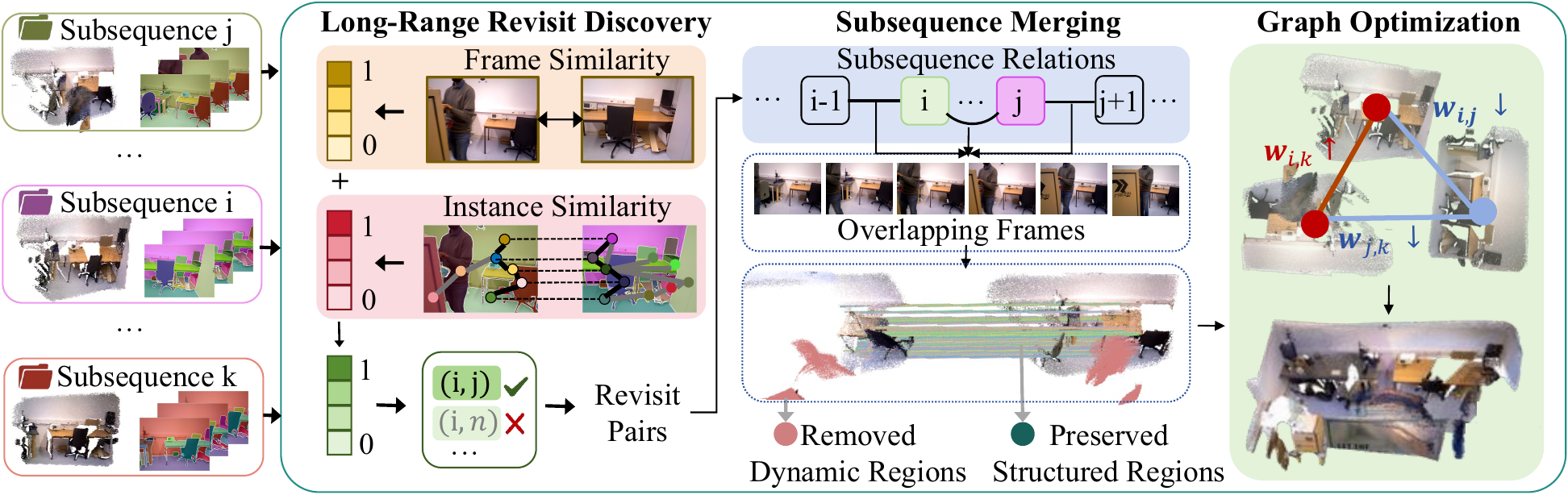}
  \vspace{-1.6em}
  \caption{
  \textbf{Instance-guided Cross-subsequence Association.}
  Given local instance libraries and per-subsequence reconstruction outputs, LIST3R establishes reliable cross-subsequence connections through long-range revisit discovery and instance-aware subsequence merging.
  The resulting adjacent and loop edges are optimized in a global association graph to obtain globally aligned subsequences.
  }
  \label{fig:method_pipeline}
  \vspace{-1.0em}
\end{figure*}

We use the local instance libraries to establish spatial associations across different subsequences.
For the $k$-th subsequence $\mathcal{S}^{(k)}$, a 3D reconstruction foundation model~\cite{wang2025pi} predicts local point maps $\mathbf{P}^{(k)}$, geometric confidence maps $\mathbf{C}^{(k)}$, and camera poses $\mathbf{T}^{(k)}$.
Based on these local geometric predictions and the instance anchors stored in each local instance library, we complete this process in three stages, as illustrated in Fig.~\ref{fig:method_pipeline}.
We first retrieve subsequences that observe revisited regions, then merge the associated subsequences through instance-aware correspondences, and finally refine the assembled scene with confidence-weighted optimization.
We describe these three components in detail below.

\paragraph{Instance-enhanced long-range revisit discovery.}
For long-range revisits, we use a frame-level loop detector to produce coarse revisit candidates with an appearance-based score $s_{\mathrm{det}}(u,v)$ followed~\cite{deng2025vggt}.
Since frame-level similarity alone is often insufficient for reliable long-range loop closure under viewpoint changes and occlusions, we use it only as a candidate proposal stage.
We then strengthen each candidate revisit using the local instance library from two complementary aspects: object-level semantic consistency and spatial relation consistency.

For a candidate pair $(u,v)$, we first match local instances across the two subsequences using their aggregated instance representations.
Given two local instance anchors $\ell_i^{(u)}\in\mathcal{L}_{\mathrm{loc}}^{(u)}$ and $\ell_j^{(v)}\in\mathcal{L}_{\mathrm{loc}}^{(v)}$, we compute the cosine similarity between $\mathbf{z}_i^{(u)}$ and $\mathbf{z}_j^{(v)}$.
We then apply mutual nearest matching in this feature space to obtain candidate instance correspondences $\mathcal{P}_{uv}$.
The semantic consistency score $s_{\mathrm{sem}}(u,v)$ is computed as the average similarity of the matched pairs, measuring whether the two subsequences contain similar object instances.

However, semantic similarity alone may be ambiguous in scenes with repeated objects, such as multiple chairs, cabinets, or tables.
To reduce this ambiguity, we further compare the relative spatial layout of the matched instances.
For each matched instance, we estimate its 3D center from the masked local geometry.
Given the matched instance correspondences $\mathcal{P}_{uv}$, we compare the relative spatial relations among the matched instances in the two subsequences.
This yields an instance spatial relation score $s_{\mathrm{rel}}(u,v)$, measuring whether matched objects preserve a consistent relative layout.

The final long-range revisit score combines frame-level appearance similarity, instance-level semantic consistency, and spatial layout consistency:
\begin{equation}
s_{\mathrm{loop}}(u,v)
=
\lambda s_{\mathrm{det}}(u,v)
+
(1-\lambda)
\left(
s_{\mathrm{sem}}(u,v)
+
s_{\mathrm{rel}}(u,v)
\right).
\label{revisit_score}
\end{equation}
We select reliable revisited frame pairs according to $s_{\mathrm{loop}}(u,v)$, and use their corresponding subsequences as loop candidates for subsequent merging.

\paragraph{Instance-aware subsequence merging.}
During subsequence merging, reliable correspondences need to be selected from two local reconstructions to estimate their relative transformation.
A straightforward strategy is to select high-confidence points according to geometric confidence~\cite{deng2025vggt}.
Although this can reduce the influence of dynamic objects to some extent, it may also discard static objects with strong structural value~\cite{wang2025vggt}.
For example, tables, chairs, and cabinets may receive low confidence under large viewpoint changes, occlusions, or incomplete local reconstruction.
If only high-confidence points are retained, the registration support often concentrates on large planar regions such as walls and floors, while object regions that provide useful structural constraints are removed.
This can make subsequence connection rigid or inaccurate.

For two overlapping subsequences $\mathcal{S}^{(u)}$ and $\mathcal{S}^{(v)}$, we first select confidence-supported candidate regions from the local point maps and confidence maps.
We then evaluate the motion consistency of each instance region in the overlap.
For an instance region $o$, we define its motion inconsistency as
\begin{equation}
\delta(o)
=
\frac{1}{|\Omega_o|}
\sum_{p\in\Omega_o}
\left\|
\pi\!\left(
\widehat{\mathbf{T}}_{uv}\mathbf{P}^{(u)}(p)
\right)
-
\phi_{uv}(p)
\right\|_2 ,
\end{equation}
where $\Omega_o$ denotes the mask region of instance $o$, $\mathbf{P}^{(u)}(p)$ is the local 3D point at pixel $p$, $\widehat{\mathbf{T}}_{uv}$ is the current relative transformation from subsequence $u$ to subsequence $v$, $\pi(\cdot)$ is the projection function, and $\phi_{uv}(p)$ denotes the observed correspondence of $p$ in subsequence $v$.

A large $\delta(o)$ indicates that the projected geometry of this instance is inconsistent with the observed correspondence, suggesting that the region is dynamic, incorrectly tracked, or locally unstable.
We remove such regions from the candidate support.
From the remaining regions, we sample the final alignment support by considering both geometric confidence and spatial coverage.
This allows the alignment support to retain structurally informative static objects, rather than being dominated by large planar regions, leading to more stable and accurate subsequence merging.

\paragraph{Cross-subsequence graph optimization.}
After the above steps, we obtain a set of pairwise subsequence relations, including adjacent relations induced by temporal overlap and loop relations discovered from revisited regions.
We formulate these relations as a global association graph~\cite{agarwal2013robust,sunderhauf2012switchable}, where each subsequence is a node and each pairwise relation is an edge.
For every connected pair $(u,v)$, we apply the same instance-aware merging to select reliable correspondences between the two local point clouds, and estimate a relative Sim(3) transformation $\widehat{\mathbf{T}}_{uv}$ from these correspondences~\cite{strasdat2010scale,umeyama2002least}.
Each edge is assigned a confidence weight $w_{uv}$, computed from the average confidence of the selected alignment points.
Let $\mathcal{E}$ denote the resulting edge set, including both adjacent edges and instance-enhanced loop edges.
We optimize the global transformation $\mathbf{G}_k\in\mathrm{Sim}(3)$ of each subsequence by minimizing
\begin{equation}
\{\mathbf{G}_k^\star\}
=
\arg\min_{\{\mathbf{G}_k\}}
\sum_{(u,v)\in\mathcal{E}}
w_{uv}
\left\|
\log\left(
\widehat{\mathbf{T}}_{uv}^{-1}
\mathbf{G}_v^{-1}
\mathbf{G}_u
\right)
\right\|_2^2 .
\end{equation}
After optimization, the aligned subsequences are assembled into a complete scene reconstruction.

\subsection{Global Instance Library}
\label{sec:global_instance_library}

\begin{figure*}[t]
  \centering
  \includegraphics[width=\textwidth]{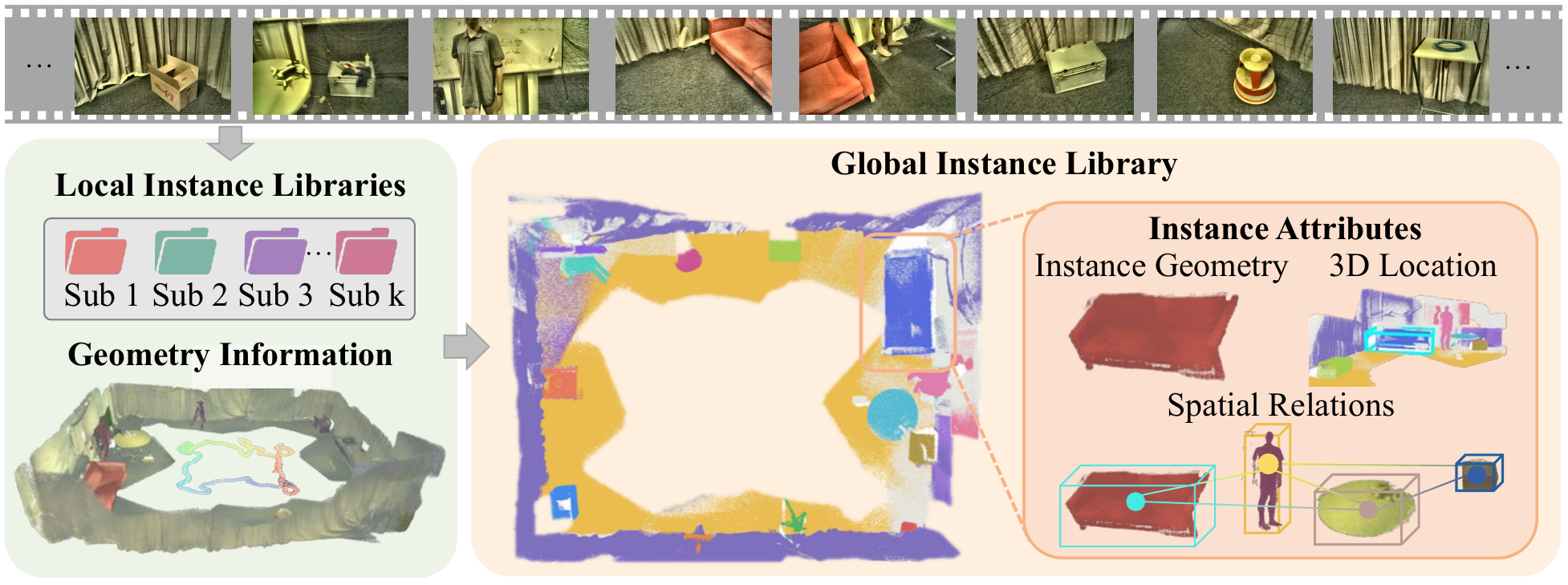}
  \vspace{-1.5em}
  \caption{
  \textbf{Global Instance Library.}
  Based on local instance libraries and reconstructed geometry, LIST3R updates instance evidence and organizes local instance records into a global 3D instance library.
  Each global instance record maintains object-level geometry, global location, and spatial relations, providing a persistent scene-level representation.
  }
  \label{fig:global_instance_library}
  \vspace{-1.5em}
\end{figure*}

To build a structured object-level representation of the reconstructed scene, we consolidate local instances into a global 3D instance library, as illustrated in Fig.~\ref{fig:global_instance_library}.
After cross-subsequence alignment in Sec.~\ref{sec:instance_guided_association}, each local instance has a spatial location in the global reference frame, while its identity is still defined locally within its subsequence.
We therefore merge local instances that correspond to the same physical object, producing persistent global instance records across the whole sequence.

For each local instance, we extract its 3D support from the globally aligned point maps according to its temporally consistent 2D masks.
This converts local mask tracks into local 3D instance nodes, where each node contains both its 2D observations and its global geometric support.
We then compare local nodes from different subsequences globally.
Nodes with overlapping global 3D support and compatible instance features are connected by merge edges.
Connected local nodes are further merged into a unified global instance record.
The global 3D instance library is defined as
\begin{equation}
\begin{aligned}
\mathcal{L}_{\mathrm{glob}}
&=
\left\{
g_m
\right\}_{m=1}^{M},
\qquad
g_m
=
\left(
\mathrm{id}_m,
\mathcal{O}_m,
\mathcal{G}_m,
\mathcal{D}_m
\right),
\\
\mathcal{O}_m
&=
\left\{
(k,i,\ell_i^{(k)})
\mid
\ell_i^{(k)} \mapsto g_m
\right\},
\\
\mathcal{G}_m
&=
\left(
\mathbf{X}_m,
\mathbf{C}_m
\right),
\qquad
\mathcal{D}_m
=
\left(
\mathbf{c}_m,
\mathbf{b}_m,
\bar{\mathbf{z}}_m,
\mathcal{R}_m
\right).
\end{aligned}
\end{equation}
where $\mathrm{id}_m$ denotes the unified global instance ID.
$\mathcal{O}_m$ denotes the set of local instance observations assigned to the $m$-th global instance, where each $\ell_i^{(k)}$ preserves its local mask track, visible frames, 2D locations, and local feature.
$\mathcal{G}_m$ denotes the aggregated global geometric evidence, including the global 3D point support $\mathbf{X}_m$ and its associated confidence $\mathbf{C}_m$.
$\mathcal{D}_m$ denotes the derived global instance attributes, including the global 3D location $\mathbf{c}_m$, spatial position $\mathbf{b}_m$, aggregated instance feature $\bar{\mathbf{z}}_m$, and object-level spatial relations $\mathcal{R}_m$.

Through this consolidation, local fragments of the same physical object observed in different subsequences are grouped into a single global instance.
This library provides a structured object-level scene representation for long-sequence reconstruction, and can serve as a 3D memory basis for downstream VLMs to perform object querying, spatial relation understanding, and scene-level reasoning.

\section{Experiments}
\label{sec:exps}

\subsection{Experimental Setup}\label{sec:exp_setup}

We evaluate LIST3R on two complementary long-sequence 3D reconstruction tasks:
camera trajectory estimation and point cloud reconstruction.
The former measures whether a method can maintain globally consistent camera motion over long videos, while the latter evaluates whether the fused geometry remains accurate and complete after cross-subsequence integration.
All experiments are conducted on a single NVIDIA RTX 4090 GPU.
For fair comparison, all methods are evaluated on the same input sequences and under the same metric protocols.

\paragraph{\textbf{Datasets and metrics.}}
For camera trajectory evaluation, we use TUM~\cite{sturm2012benchmark}, ETH3D~\cite{schops2019bad}, and BONN~\cite{palazzolo2019refusion}, which cover indoor sequences, challenging camera motion, and long-range revisits. 
We report ATE, RTE, and RRE to evaluate both global trajectory accuracy and accumulated drift~\cite{sturm2012benchmark,wang2025continuous,deng2025vggt}. 
For point cloud reconstruction, we evaluate on ETH3D~\cite{schops2019bad} and NRGBD~\cite{azinovic2022neural}. 
ETH3D contains larger-scale scenes with challenging viewpoint changes and diverse trajectories, while NRGBD provides sequences with dense ground-truth geometry for fine-grained reconstruction evaluation.
We report Chamfer distance, Accuracy, Completion, Normal Consistency (NC), and F-score at 5cm, following prior works~\cite{schops2017multi,wang2025continuous,deng2025vggt}. 
Together, these benchmarks evaluate whether LIST3R improves long-sequence reconstruction consistency while preserving local geometric accuracy.

\paragraph{\textbf{Baselines.}}
We compare LIST3R with several methods.
VGGT~\cite{wang2025vggt} and $\pi^3$~\cite{wang2025pi} are included as direct feed-forward baselines.
For scalable long-sequence reconstruction, we compare with CUT3R~\cite{wang2025continuous}, TTT3R~\cite{chen2025ttt3r}, VGGT-Long~\cite{deng2025vggt}, and Scal3R~\cite{xie2026scal3r}.
In addition, we introduce a strong subsequence-based baseline, denoted as $\pi$-Long, which follows the standard split-and-fuse configuration for long-sequence reconstruction and uses $\pi^3$ as the base reconstruction model.

\begin{table}[t]
\caption{
\textbf{Camera pose estimation on long sequences.}
We report ATE, RTE, and RRE.
All metrics are lower better.
OOM indicates out-of-memory when directly processing the full sequence.
}
\label{tab:camera_pose}
\centering
\renewcommand{\arraystretch}{1.02}
\renewcommand{\tabcolsep}{1.5pt}
\resizebox{\linewidth}{!}{
\begin{tabular}{@{}llccc|ccc|ccc@{}}
\toprule
 &    &
 \multicolumn{3}{c}{\textbf{TUM~\cite{sturm2012benchmark}}}
 & \multicolumn{3}{c}{\textbf{ETH3D~\cite{schops2019bad}}}
 & \multicolumn{3}{c}{\textbf{BONN~\cite{palazzolo2019refusion}}} \\
\cmidrule(lr){3-5} \cmidrule(lr){6-8} \cmidrule(lr){9-11}
 & \textbf{Method}
 & {ATE (m) $\downarrow$} & {RTE (m) $\downarrow$} & {RRE (deg) $\downarrow$}
 & {ATE (m) $\downarrow$} & {RTE (m) $\downarrow$} & {RRE (deg) $\downarrow$}
 & {ATE (m) $\downarrow$} & {RTE (m) $\downarrow$} & {RRE (deg) $\downarrow$} \\
\midrule

 & \cg VGGT~\cite{wang2025vggt}
 & \cg OOM & \cg OOM & \cg OOM
 & \cg OOM & \cg OOM & \cg OOM
 & \cg OOM & \cg OOM & \cg OOM \\

 & \cg $\pi^3$~\cite{wang2025pi}
 & \cg OOM & \cg OOM & \cg OOM
 & \cg OOM & \cg OOM & \cg OOM
 & \cg OOM & \cg OOM & \cg OOM \\

\midrule
 & CUT3R~\cite{wang2025continuous}
 & 0.866 & 0.963 & 40.189
 & 2.895 & 2.537 & 43.040
 & 0.319 & \firstc{\textbf{0.561}} & 58.127 \\

 & TTT3R~\cite{chen2025ttt3r}
 & 0.317 & 0.385 & 9.923
 & 1.317 & 0.939 & 10.326
 & 0.149 & \secondc{0.759} & 47.509 \\

 & VGGT-Long~\cite{deng2025vggt}
 & 0.325 & 0.489 & 25.209
 & 1.292 & 1.701 & 32.922
 & 0.123 & 0.787 & \secondc{47.426} \\

 & $\pi$-Long
 & \secondc{0.208} & \secondc{0.279} & 7.805
 & \secondc{0.562} & \secondc{0.455} & 13.654
 & \secondc{0.094} & 0.770 & 48.008 \\

 & Scal3R~\cite{xie2026scal3r}
 & 0.267 & 0.329 & \firstc{\textbf{5.724}}
 & 0.807 & 0.590 & \firstc{\textbf{7.002}}
 & 0.117 & 0.779 & 49.093 \\

 & \textbf{Ours}
 & \firstc{\textbf{0.150}} & \firstc{\textbf{0.211}} & \secondc{6.974}
 & \firstc{\textbf{0.516}} & \firstc{\textbf{0.444}} & \secondc{9.322}
 & \firstc{\textbf{0.085}} & 0.779 & \firstc{\textbf{45.890}} \\

\bottomrule
\end{tabular}
}
\end{table}

\begin{figure}[t]
  \centering
  \includegraphics[width=\linewidth]{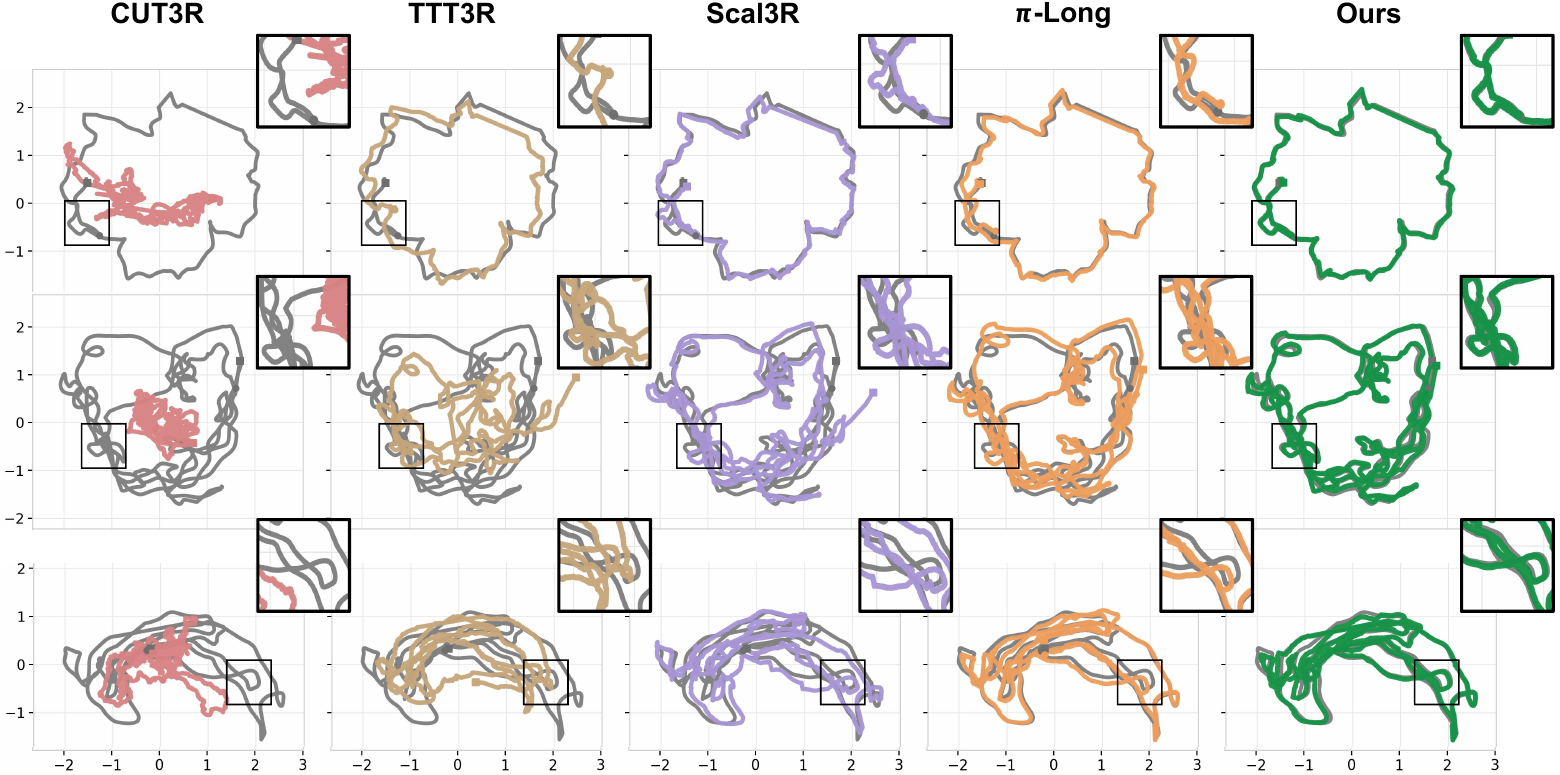}
  \caption{Visualization of Estimated Long-Sequence Camera Trajectories.}
  \label{fig:camera_pose}
\end{figure}

\begin{table}[t]
\caption{
\textbf{Point cloud reconstruction quality.}
We report reconstruction quality on ETH3D and NRGBD using Chamfer distance, accuracy, completeness, normal consistency, and F-score at 5cm.
}
\label{tab:point_cloud_quality}
\centering
\renewcommand{\arraystretch}{1.02}
\renewcommand{\tabcolsep}{3pt}
\resizebox{\linewidth}{!}{
\begin{tabular}{@{}llccccc|ccccc@{}}
\toprule
 &    &
 \multicolumn{5}{c}{\textbf{ETH3D~\cite{schops2019bad}}}
 & \multicolumn{5}{c}{\textbf{NRGBD~\cite{azinovic2022neural}}} \\
\cmidrule(lr){3-7} \cmidrule(lr){8-12}
 & \textbf{Method}
 & {Chamfer $\downarrow$} & {Acc. $\downarrow$} & {Comp. $\downarrow$} & {NC $\uparrow$} & {F@5cm $\uparrow$}
 & {Chamfer $\downarrow$} & {Acc. $\downarrow$} & {Comp. $\downarrow$} & {NC $\uparrow$} & {F@5cm $\uparrow$} \\
\midrule

 & \cg VGGT~\cite{wang2025vggt}
 & \cg OOM & \cg OOM & \cg OOM & \cg OOM & \cg OOM
 & \cg OOM & \cg OOM & \cg OOM & \cg OOM & \cg OOM \\

 & \cg $\pi^3$~\cite{wang2025pi}
 & \cg OOM & \cg OOM & \cg OOM & \cg OOM & \cg OOM
 & \cg OOM & \cg OOM & \cg OOM & \cg OOM & \cg OOM \\

\midrule
 & CUT3R~\cite{wang2025continuous}
 & 140.049 & 62.803 & 217.294 & 0.536 & 3.995
 & 73.239 & 50.051 & 96.427 & 0.575 & 9.009 \\

 & TTT3R~\cite{chen2025ttt3r}
 & 102.590 & 36.604 & 168.575 & 0.610 & 7.274
 & 41.287 & 26.162 & 56.413 & 0.647 & 22.236 \\

 & VGGT-Long~\cite{deng2025vggt}
 & 50.551 & 56.748 & 44.354 & 0.618 & 19.834
 & 6.097 & 5.334 & 6.860 & 0.857 & 68.878 \\

 & $\pi$-Long
 & 41.536 & 37.775 & 45.296 & \secondc{0.686} & \secondc{32.655}
 & \secondc{4.973} & 4.412 & \secondc{5.535} & \firstc{\textbf{0.876}} & 68.917 \\

 & Scal3R~\cite{xie2026scal3r}
 & \secondc{33.790} & \secondc{36.451} & \secondc{31.129} & 0.658 & 26.203
 & 7.669 & \secondc{4.231} & 11.106 & 0.829 & \secondc{71.203} \\

 & \textbf{Ours}
 & \firstc{\textbf{27.362}} & \firstc{\textbf{31.094}} & \firstc{\textbf{23.629}} & \firstc{\textbf{0.709}} & \firstc{\textbf{36.450}}
 & \firstc{\textbf{4.680}} & \firstc{\textbf{4.104}} & \firstc{\textbf{5.256}} & \secondc{0.875} & \firstc{\textbf{73.363}} \\

\bottomrule
\end{tabular}
}
\vspace{-9mm}
\begin{flushleft}
\end{flushleft}
\end{table}

\begin{figure}[t]
  \centering
  \includegraphics[width=\linewidth]{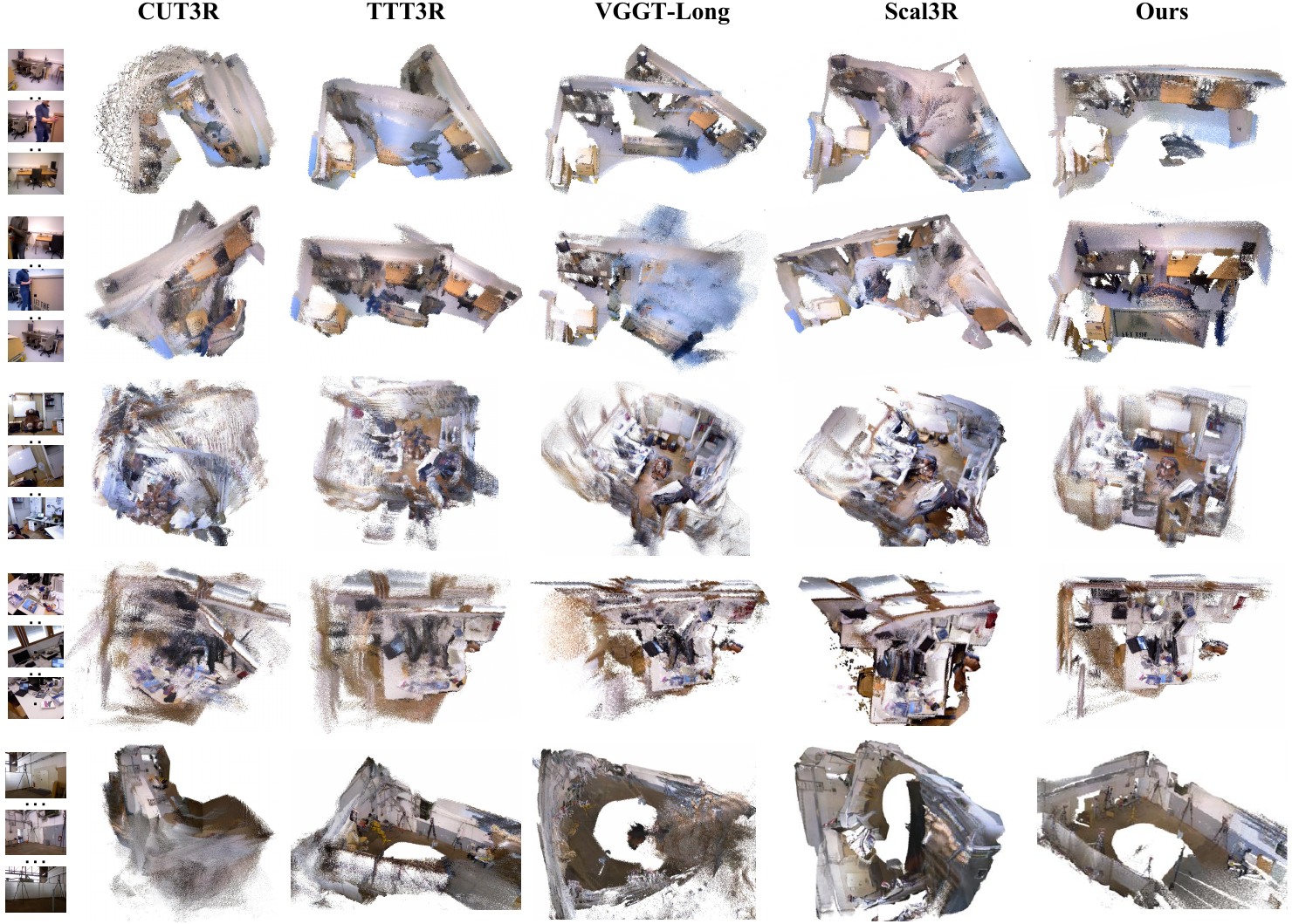}
  \caption{\textbf{Qualitative results for long-sequence 3D reconstruction.}}
  \vspace{-10pt}
  \label{fig:pose_vis_single}
\end{figure}

\subsection{Trajectory Evaluation}
\label{sec:traj_eval}

We evaluate camera pose accuracy on TUM, ETH3D, and BONN, with results reported in Table~\ref{tab:camera_pose}.
Directly applying VGGT and $\pi^3$ to full long sequences leads to out-of-memory errors, highlighting the need for scalable long-sequence reconstruction.
This setting is particularly challenging because the input video must be decomposed into local subsequences, while the final trajectory still needs to remain globally consistent across long temporal gaps, viewpoint changes, and possible revisits.

Under this setting, LIST3R achieves the best ATE on all three datasets, demonstrating stronger global trajectory consistency.
Since ATE measures the error of the entire aligned camera path, it directly reflects whether local subsequences are correctly connected into a coherent global trajectory.
On TUM and ETH3D, LIST3R also achieves the best RTE, showing that the recovered long-range constraints further improve relative translation consistency after subsequence fusion.

For RTE and RRE, some baselines can still be favored by local trajectory characteristics.
These metrics evaluate relative motion over local trajectory segments, and thus can benefit methods that preserve short-term smoothness or local rotational consistency.
For example, Scal3R obtains lower RRE on TUM and ETH3D, while CUT3R and TTT3R achieve lower RTE on BONN.
In LIST3R, the optimization prioritizes globally reliable subsequence relations: weak or low-confidence cross-subsequence connections are assigned lower weights, since local reconstructions and overlap regions can be noisy or poorly constrained.
As a result, some local relative errors may increase around weak overlaps between subsequences, but the optimized trajectory better satisfies high-confidence long-range constraints and achieves stronger global consistency.
This trade-off is reflected by the best ATE across all datasets, while the relative pose metrics remain competitive with strong baselines.
The qualitative trajectories in Fig.~\ref{fig:camera_pose} further support this conclusion.
Compared with baselines, LIST3R follows the reference trajectory more closely, especially in regions where accumulated drift becomes visible.
Overall, these results demonstrate the effectiveness of persistent instance anchors for converting fragmented local reconstructions into a globally consistent trajectory.

\vspace{-3pt}
\subsection{Geometry Evaluation}\label{sec:geo_eval}

We further evaluate point cloud reconstruction quality on ETH3D and NRGBD, as summarized in Table~\ref{tab:point_cloud_quality}.
On ETH3D, LIST3R achieves the best performance across all five metrics, reducing Chamfer distance, accuracy error, and completeness error while obtaining higher normal consistency and F@5cm.
This indicates that the proposed instance-guided associations not only improve cross-subsequence consistency, but also lead to more complete and geometrically coherent point clouds after long-sequence fusion.
On NRGBD, LIST3R also obtains the best Chamfer, accuracy, completeness, and F@5cm, while achieving normal consistency comparable to the strongest baseline.
Compared with $\pi$-Long, LIST3R improves Chamfer from 4.973 to 4.680 and F@5cm from 68.917 to 73.363, suggesting that the recovered long-range constraints help reduce accumulated misalignment and improve the final fused geometry.
Compared with Scal3R, LIST3R is substantially better in completeness and Chamfer, showing that object-level anchors provide more reliable evidence for aligning revisited regions than purely geometric matching.

Overall, these results demonstrate that LIST3R improves global reconstruction consistency without sacrificing local geometric quality.
The gains are especially clear in completeness and F-score, which are directly affected by whether different subsequences are correctly merged into a coherent scene.
Qualitative comparisons are provided in Fig.~\ref{fig:pose_vis_single}.

\subsection{Ablation Study}
\label{sec:ablation}

We conduct an ablation study on LIST3R to analyze its core module,
Instance-guided Cross-subsequence Association.
Starting from the $\pi$-Long baseline, which follows the standard split-and-fuse long-sequence reconstruction pipeline using $\pi^3$~\cite{wang2025pi} as the base model, we progressively enable three components:
(1) Instance-enhanced Long-range Revisit Discovery (ILRD), which retrieves cross-subsequence loop candidates with instance cues;
(2) Instance-aware Subsequence Merging (IASM), which uses instance correspondences to improve subsequence alignment;
and (3) Confidence-weighted Cross-subsequence Graph Optimization (CWGO), which optimizes the global subsequence graph with confidence-aware constraints.

As shown in Table~\ref{tab:ablation_pose} and Table~\ref{tab:ablation_recon}, the Baseline suffers from both large trajectory drift and incomplete reconstruction.
Adding ILRD brings clear improvements in pose estimation and point cloud quality, showing that instance-enhanced revisit discovery recovers useful long-range constraints across fragmented subsequences.
IASM further improves the results by filtering unreliable cross-subsequence associations with instance-level correspondences and dynamic-instance cues.
Finally, CWGO achieves the best results on all metrics, confirming that it effectively turns instance-guided constraints into globally consistent poses and point clouds.
Overall, the ablation confirms that the three components are complementary: ILRD expands long-range associations, IASM improves their reliability, and CWGO consolidates them into a coherent global reconstruction.

\begin{table}[t]
\centering

\begin{minipage}[t]{0.48\linewidth}
\centering
\caption{
\textbf{Camera pose ablation.}
}
\label{tab:ablation_pose}
\renewcommand{\arraystretch}{1.08}
\setlength{\tabcolsep}{3pt}
\resizebox{\linewidth}{!}{
\begin{tabular}{@{}lccc@{}}
\toprule
\textbf{Variant} &
\textbf{ATE (m) $\downarrow$} &
\textbf{RTE (m) $\downarrow$} &
\textbf{RRE (deg) $\downarrow$} \\
\midrule
Baseline
& 0.964 & 1.231 & 17.908 \\

+ ILRD
& 0.708 & 0.724 & 13.351 \\

+ IASM
& 0.706 & 0.720 & 13.340 \\

+ CWGO (Full)
& \textbf{0.277} & \textbf{0.402} & \textbf{6.920} \\

\bottomrule
\end{tabular}
}
\end{minipage}
\hfill
\begin{minipage}[t]{0.48\linewidth}
\centering
\caption{
\textbf{Reconstruction quality ablation.}
}
\label{tab:ablation_recon}
\renewcommand{\arraystretch}{1.08}
\setlength{\tabcolsep}{6pt}
\resizebox{\linewidth}{!}{
\begin{tabular}{@{}lccc@{}}
\toprule
\textbf{Variant} &
\textbf{Comp. $\downarrow$} &
\textbf{Chamfer $\downarrow$} &
\textbf{F@5cm $\uparrow$} \\
\midrule
Baseline
& 168.246 & 149.726 & 2.29 \\

+ ILRD
& 86.192 & 125.382 & 4.80 \\

+ IASM
& 82.961 & 121.684 & 4.91 \\

+ CWGO (Full)
& \textbf{32.719} & \textbf{35.325} & \textbf{9.91} \\

\bottomrule
\end{tabular}
}

\end{minipage}
\vspace{-4mm}
\end{table}
\section{Conclusion}
\label{sec:conclusion}

We presented LIST3R, an instance-aware framework for long-sequence 3D reconstruction that organizes fragmented subsequences around persistent instance anchors.
LIST3R builds local instance libraries within partial reconstructions and uses the stored instance evidence to recover revisited regions and guide object-aware fragment alignment.
As reconstruction proceeds, the updated geometric evidence further refines the local libraries and progressively forms a unified global 3D instance library, enabling a coherent scene-level representation across the full sequence.
Experiments on long-sequence benchmarks show that LIST3R produces more accurate trajectories and higher-quality 3D reconstructions than existing baselines.
These results demonstrate that persistent instance anchors provide an effective and scalable cue for organizing long-horizon 3D reconstruction.

{\small
\bibliographystyle{plain}
\bibliography{main}
}








\newpage

\appendix

\section*{Appendix} 
\vspace{1em}

\begin{center}
{\large \bfseries Table of Contents}
\end{center}
\vspace{-1.5em}

\noindent\rule{\linewidth}{0.4pt}

\vspace{0.5em}

{

\vspace{0.5em}

\noindent
\textbf{A \quad Additional Implementation Details} \dotfill \pageref{app:implementation} \\
\hspace*{1.5em} A.1 \quad Local Instance Anchor Initialization \dotfill \pageref{app:local_instance_anchor} \\
\hspace*{1.5em} A.2 \quad Long-range Revisit Discovery \dotfill \pageref{app:revisit_analysis} \\
\hspace*{1.5em} A.3 \quad Instance-aware Subsequence Merging \dotfill \pageref{app:subsequence_merging} \\
\hspace*{1.5em} A.4 \quad Cross-subsequence Graph Optimization \dotfill \pageref{app:graph_optimization} \\[0.5em]

\textbf{B \quad Additional Experimental Results} \dotfill \pageref{app:additional_results} \\
\hspace*{1.5em} B.1 \quad Additional Quantitative Results \dotfill \pageref{app:trajectory_visualizations} \\
\hspace*{1.5em} B.2 \quad Additional Camera Trajectory Visualizations \dotfill \pageref{app:reconstruction_visualizations} \\
\hspace*{1.5em} B.3 \quad Additional Point Cloud Reconstruction Visualizations \dotfill \pageref{app:reconstruction_visualizations} \\[0.5em]

\textbf{C \quad Limitations} \dotfill \pageref{app:limitations} \\[0.5em]

\textbf{D \quad Broader Impacts} \dotfill \pageref{app:broader_impacts}

}

\vspace{0.5em}

\noindent\rule{\linewidth}{0.4pt}

\vspace{1em}

\section{Additional Implementation Details}
\label{app:implementation} 

\subsection{Local Instance Anchor Initialization}
\label{app:local_instance_anchor}

We provide additional implementation details for constructing the local instance libraries described in Sec.~\ref{sec:local_instance_initialization}.
To keep the local instance initialization lightweight while maintaining coverage of different viewpoints, we sample one anchor frame every 50 frames.
Accordingly, each subsequence is set to 100 frames, so that two anchor frames are selected within each subsequence.
Following VGGT-Long~\cite{deng2025vggt}, adjacent subsequences are constructed with a 30-frame overlap to preserve boundary observations and support cross-subsequence association.
The overlap between adjacent subsequences further helps preserve boundary observations and facilitates later cross-subsequence association.

The sampled anchor frames are used for query-based instance discovery.
For each anchor frame, the instance discovery module predicts candidate masks together with their object-level features.
These candidates serve as the initial source of local instance anchors, but directly using all predicted masks can introduce noisy, fragmented, or redundant object observations.
We therefore perform a simple prompt selection step before temporal propagation.
Specifically, candidate masks are filtered according to mask confidence, mask area, and mask redundancy.
Low-confidence masks and very small masks are removed, while highly overlapping duplicate masks are suppressed.
This filtering step keeps the initialization compact and reduces the chance that unstable masks are propagated to the full subsequence.

The remaining masks are then used as initialization prompts for SAM3~\cite{carion2025sam} propagation over the whole subsequence.
After propagation, we further retain only instance tracks that are temporally stable and sufficiently observed within the subsequence.
For each retained local instance, its propagated multi-frame masks are collected as the mask track $\mathcal{M}_i^{(k)}$.
The corresponding instance feature $\mathbf{z}_i^{(k)}$ is obtained from the object features of its associated initialized candidates.
In our implementation, we first normalize these candidate features and then average them to obtain a compact object-level descriptor.
This produces a local instance record that jointly contains temporally consistent masks and an aggregated object feature, which are later used for cross-subsequence association.

\begin{figure*}[h]
  \centering
  \includegraphics[width=0.95\textwidth]{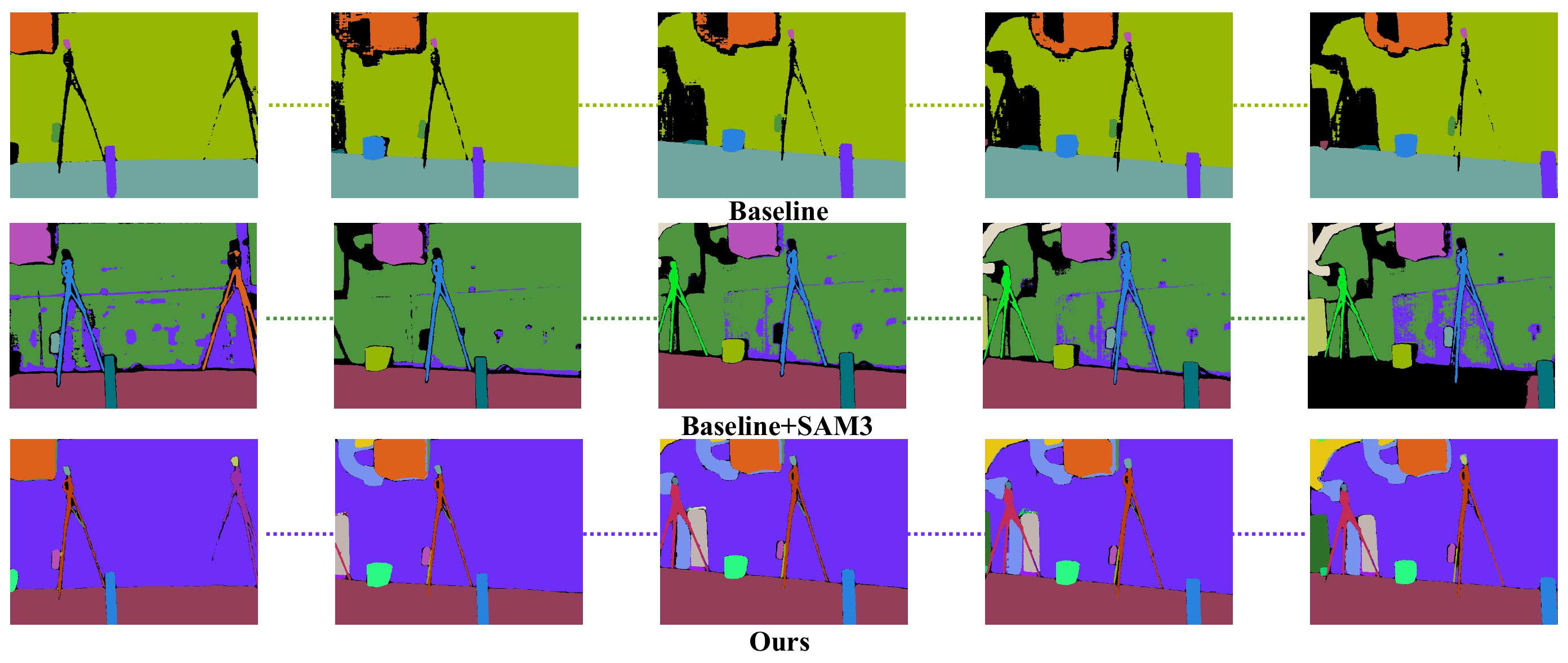}
  \vspace{-0.6em}
  \caption{
  \textbf{Comparison of local instance anchor initialization strategies.}
  We compare three strategies for constructing local instance anchors across nearby frames: directly using EntitySAM predictions, directly propagating all discovered masks with SAM3, and our final anchor-enhanced initialization with prompt selection.
  From top to bottom, the three rows show the results of these strategies, respectively.
  }
  \label{fig:entitysam3}
  \vspace{-1.0em}
\end{figure*}

Fig.~\ref{fig:entitysam3} provides a qualitative comparison of different local instance anchor initialization strategies.
Directly using instance discovery results can produce fragmented or inconsistent masks across nearby frames.
Directly injecting all discovered masks into SAM3 improves temporal coverage, but noisy initial prompts may still lead to unstable propagation.
In contrast, our final strategy selects reliable prompts before propagation, leading to cleaner mask tracks and more continuous local instance anchors.
These stable local anchors provide more reliable evidence for subsequent revisit discovery and subsequence merging.

\subsection{Long-range Revisit Discovery}
\label{app:revisit_analysis}

We provide additional analysis of the long-range revisit discovery stage.
Existing subsequence-based methods commonly rely on frame-level similarity for loop detection~\cite{deng2025vggt}. 
While such a criterion is effective for visually similar revisits, it can miss important long-range connections when the revisited region is observed under large viewpoint changes, partial occlusion, or different object visibility.
In addition, nearby or overlapping subsequences often produce high frame-level similarity scores and occupy the top-ranked candidates, although these local connections contribute less to discovering long-range revisits.
Simply lowering the loop threshold can improve recall, but it also risks introducing unreliable low-confidence loop edges into the global optimization.

To address this issue, LIST3R decouples loop proposal generation from loop verification.
We first use frame-level similarity as a recall-oriented preselection cue.
Since adjacent subsequences are already connected through the overlap setting, we exclude neighboring subsequences from the long-range loop candidate set and use a relaxed frame-level threshold of 0.5 to retain potential revisits.
This loose preselection is designed to avoid discarding true long-range loops at an early stage.
The retained candidates are then verified using instance-level evidence from the local instance libraries constructed in Sec.~\ref{sec:local_instance_initialization}.
Specifically, we compute an instance semantic consistency score from the cosine similarity between local instance features $\mathbf{z}_i^{(k)}$, and further combine it with an instance spatial relation consistency score.
The final loop confidence is computed according to Eq.~\ref{revisit_score}.

\begin{figure*}[t]
  \centering
  \includegraphics[width=0.95\textwidth]{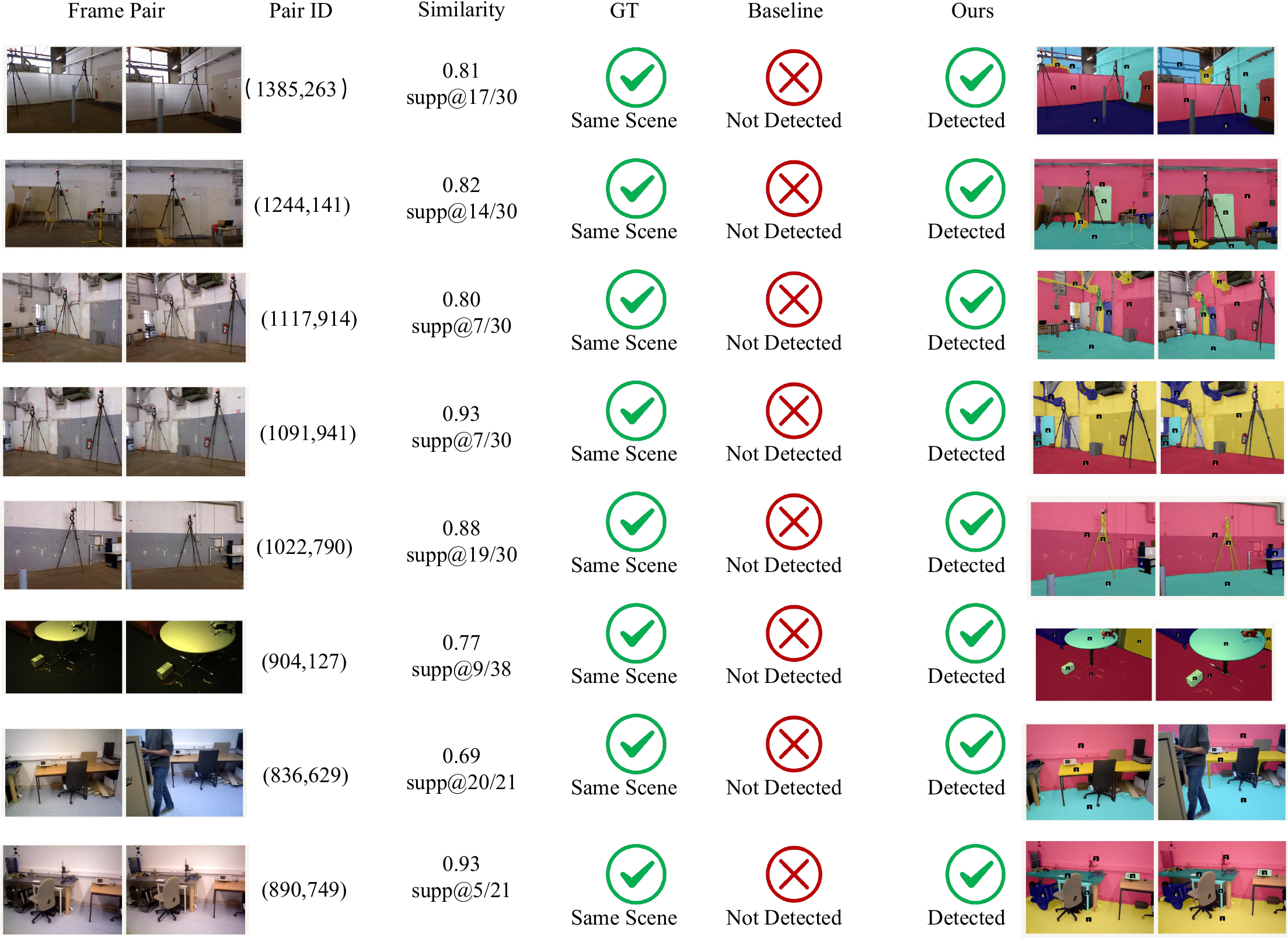}
  \vspace{-0.6em}
  \caption{
\textbf{Representative long-range revisit discovery results.}
We compare LIST3R with the baseline frame-level loop detection strategy on several challenging revisit pairs.
For each pair, we report the frame-pair ID, frame-level similarity, ground-truth relation, and whether the revisit is detected by the baseline and LIST3R.
Although these pairs correspond to the same scene, the baseline can miss them due to viewpoint changes, occlusion, or competition from nearby high-similarity frames.
The rightmost visualizations show the local instance library results of LIST3R, which provide instance-level evidence for recovering these valid long-range revisits.
}
  \label{fig:revisit}
  \vspace{-1.0em}
\end{figure*}

Fig.~\ref{fig:revisit} shows representative long-range revisit discovery results.
The baseline based on frame-level similarity fails to detect several true revisits, as these pairs are not ranked sufficiently high or have relatively weak visual similarity under viewpoint changes and occlusion.
In contrast, LIST3R recovers these valid loop connections by incorporating instance semantic consistency and spatial relation consistency.
These examples show that instance-level anchors help discover useful long-range revisits that are difficult to identify using frame-level similarity alone, providing more reliable constraints for global subsequence association.

\subsection{Instance-aware Subsequence Merging}
\label{app:subsequence_merging}

We further visualize the alignment supports used during subsequence merging.
Specifically, we compare the original confidence-based alignment strategy with our instance-aware merging strategy on representative examples.

\begin{figure*}[t]
  \centering
  \includegraphics[width=1\textwidth]{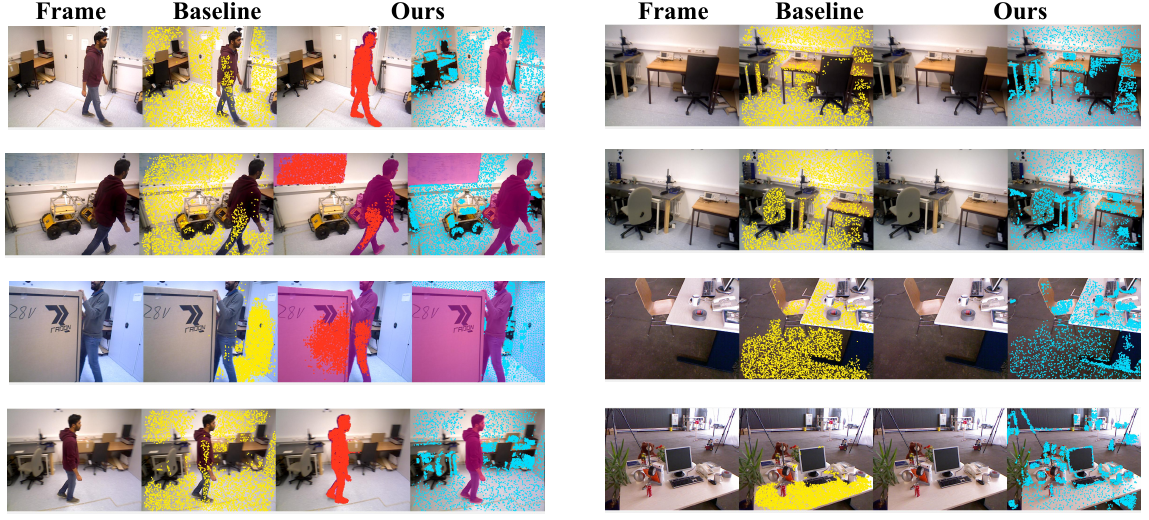}
  \vspace{-0.6em}
  \caption{
\textbf{Explanation  of instance-aware subsequence merging.}
The left block shows scenes with dynamic objects, while the right block shows mostly static scenes.
In each block, the columns show the overlapping frame, the alignment supports selected by the baseline, and the results of LIST3R.
Yellow points show the supports originally selected by the baseline, red regions mark dynamic points that are associated with the corresponding instances and removed, and cyan points indicate the final supports used by LIST3R for alignment.
LIST3R removes unstable dynamic regions when they appear, and otherwise shifts the alignment supports from large planar areas toward more structurally informative instances.
}
  \label{fig:supp_merge}
  \vspace{-1.0em}
\end{figure*}

Fig.~\ref{fig:supp_merge} Left shows cases with dynamic objects.
The original strategy often places many alignment supports on moving people or other unstable regions, which can interfere with subsequence registration.
In contrast, LIST3R identifies such dynamic instance regions and removes them from the alignment process, leading to cleaner and more reliable supports.

Fig.~\ref{fig:supp_merge} Right shows additional cases without obvious dynamic objects.
In these scenes, LIST3R does not enforce unnecessary filtering.
Instead, it plays a more important role in redistributing alignment supports: rather than being dominated by large planar regions such as floors and walls, the selected supports are shifted toward more structurally informative object instances.
This produces more spatially distributed and more reliable alignment evidence for subsequence registration.

\subsection{Cross-subsequence Graph Optimization}
\label{app:graph_optimization}

We provide additional details for the construction and optimization of the cross-subsequence association graph.
The graph contains two types of edges: adjacent edges and long-range loop edges.
Adjacent edges connect temporally neighboring subsequences, while loop edges are selected according to the instance-enhanced revisit score.
Formally, the edge set is defined as
\begin{equation}
\mathcal{E}
=
\mathcal{E}_{\mathrm{adj}}
\cup
\mathcal{E}_{\mathrm{loop}},
\qquad
\mathcal{E}_{\mathrm{adj}}
=
\{(k,k+1)\}_{k=1}^{K-1}.
\end{equation}
Here, $K$ is the number of subsequences.
$\mathcal{E}_{\mathrm{loop}}$ contains the long-range loop edges selected according to the instance-enhanced revisit score $s_{\mathrm{loop}}(u,v)$.
In practice, we keep at most top-$K_{\mathrm{loop}}$ loop candidates for each subsequence to avoid adding redundant or weak loop constraints, following VGGT-Long~\cite{deng2025vggt}.

For each connected pair $(u,v)\in\mathcal{E}$, we estimate a relative Sim(3) transformation from the selected alignment support.
Let $\mathcal{C}_{uv}$ denote the point-level correspondence set extracted from the two local point maps, where each correspondence $(\mathbf{x},\mathbf{y})$ contains a point $\mathbf{x}$ in subsequence $u$ and its corresponding point $\mathbf{y}$ in subsequence $v$.
The relative transformation is estimated by weighted alignment:
\begin{equation}
\widehat{\mathbf{T}}_{uv}
=
\arg\min_{\mathbf{T}\in\mathrm{Sim}(3)}
\sum_{(\mathbf{x},\mathbf{y})\in\mathcal{C}_{uv}}
\omega_{\mathbf{x},\mathbf{y}}
\left\|
\mathbf{T}\mathbf{x}-\mathbf{y}
\right\|_2^2 ,
\end{equation}
where $\omega_{\mathbf{x},\mathbf{y}}$ is determined by the geometric confidence of the two corresponding points.
This transformation maps points from the local coordinate system of subsequence $u$ to that of subsequence $v$.
We compute an alignment confidence $m_{uv}$ from the selected correspondences and their residuals after alignment.
Specifically, we define
\begin{equation}
r_{\mathbf{x},\mathbf{y}}
=
\left\|
\widehat{\mathbf{T}}_{uv}\mathbf{x}
-
\mathbf{y}
\right\|_2 ,
\end{equation}
and compute
\begin{equation}
m_{uv}
=
\frac{
\sum_{(\mathbf{x},\mathbf{y})\in\mathcal{C}_{uv}}
\omega_{\mathbf{x},\mathbf{y}}
\exp\left(-r_{\mathbf{x},\mathbf{y}}/\sigma_r\right)
}{
\sum_{(\mathbf{x},\mathbf{y})\in\mathcal{C}_{uv}}
\omega_{\mathbf{x},\mathbf{y}}
}.
\end{equation}
This confidence is high when the selected correspondences are reliable and well explained by the estimated relative transformation.

We then map the alignment confidence to the graph edge weight.
To reduce the effect of extreme confidence values, we first apply log compression and then sharpen the separation between weak and strong edges:
\begin{equation}
u_{uv}
=
\frac{
\log(1+\alpha m_{uv})
}{
1+\log(1+\alpha m_{uv})
},
\qquad
w_{uv}
=
\frac{
u_{uv}^{\gamma}
}{
u_{uv}^{\gamma}+(1-u_{uv})^{\gamma}
}.
\end{equation}
Here, $\alpha$ controls the strength of log compression and is set to $0.30$, while $\gamma$ controls the contrast of the final edge weight and is set to $6.0$.
Edges with very low confidence are discarded before graph optimization.

Let $\mathbf{G}_k\in\mathrm{Sim}(3)$ denote the transformation from the local coordinate system of subsequence $\mathcal{S}^{(k)}$ to the global coordinate system.
We fix the first subsequence as the reference frame:
\begin{equation}
\mathbf{G}_1=\mathbf{I}.
\end{equation}
The remaining transformations are initialized by chaining adjacent relative transformations:
\begin{equation}
\mathbf{G}_{k+1}^{(0)}
=
\mathbf{G}_{k}^{(0)}
\widehat{\mathbf{T}}_{k,k+1}^{-1},
\qquad
k=1,\ldots,K-1.
\end{equation}
After initialization, we optimize all subsequence transformations by minimizing the weighted Sim(3) pose graph objective $\mathbf{G}_k^\star$ in Sec.~\ref{sec:local_instance_initialization}.The logarithm maps the Sim(3) residual to its Lie algebra, allowing the pose graph to be optimized in a vector space.
After optimization, all subsequences are aligned into a unified global coordinate system.

\section{Additional Experimental Results}
\label{app:additional_results}

In this section, we provide supplementary experimental results to further evaluate LIST3R on long-sequence 3D reconstruction.
We first report representative quantitative results for camera trajectory estimation and point cloud reconstruction, complementing the aggregate comparisons in the main paper.
We then present additional qualitative comparisons, including camera trajectory visualizations and point cloud reconstruction results.
These results provide a more detailed view of how LIST3R improves global trajectory consistency and reconstruction quality across challenging long sequences.

\begin{table*}[t]
\caption{
\textbf{Representative camera pose estimation results, Part I.}
We report representative per-scene results for camera pose estimation, comparing LIST3R with strong subsequence-based methods, including VGGT-Long, $\pi$-Long, and Scal3R.
These sequences highlight cases where LIST3R achieves competitive or superior results across all three metrics.
Best and second-best results are highlighted.
}
\label{tab:selected_camera_pose_part1}
\centering
\renewcommand{\arraystretch}{1.02}
\setlength{\tabcolsep}{2.2pt}
\resizebox{\textwidth}{!}{
\begin{tabular}{@{}lccc|lccc@{}}
\toprule
\multicolumn{4}{c|}{\textbf{TUM}: \texttt{freiburg1\_floor}} 
& \multicolumn{4}{c}{\textbf{TUM}: \texttt{freiburg1\_room}} \\
\cmidrule(lr){1-4} \cmidrule(lr){5-8}
\textbf{Method} & {ATE (m) $\downarrow$} & {RTE (m) $\downarrow$} & {RRE (deg) $\downarrow$}
& \textbf{Method} & {ATE (m) $\downarrow$} & {RTE (m) $\downarrow$} & {RRE (deg) $\downarrow$} \\
\midrule
VGGT-Long~\cite{deng2025vggt} & 0.4976 & 0.5782 & 28.1315
& VGGT-Long~\cite{deng2025vggt} & 0.5312 & 0.5864 & 31.8505 \\
$\pi$-Long & \secondc{0.1119} & \secondc{0.1434} & \secondc{4.0399}
& $\pi$-Long & \secondc{0.0962} & \secondc{0.1062} & \secondc{2.7764} \\
Scal3R~\cite{xie2026scal3r} & 0.1230 & 0.1640 & 4.9328
& Scal3R~\cite{xie2026scal3r} & 0.1713 & 0.1814 & 3.7894 \\
\textbf{Ours} & \firstc{\textbf{0.0984}} & \firstc{\textbf{0.1153}} & \firstc{\textbf{3.8552}}
& \textbf{Ours} & \firstc{\textbf{0.0838}} & \firstc{\textbf{0.0969}} & \firstc{\textbf{2.7598}} \\

\midrule
\multicolumn{4}{c|}{\textbf{TUM}: \texttt{freiburg1\_teddy}} 
& \multicolumn{4}{c}{\textbf{TUM}: \texttt{freiburg2\_360\_hemisphere}} \\
\cmidrule(lr){1-4} \cmidrule(lr){5-8}
\textbf{Method} & ATE $\downarrow$ & RTE $\downarrow$ & RRE $\downarrow$
& \textbf{Method} & ATE $\downarrow$ & RTE $\downarrow$ & RRE $\downarrow$ \\
\midrule
VGGT-Long~\cite{deng2025vggt} & \secondc{0.0704} & 0.1473 & 20.7882
& VGGT-Long~\cite{deng2025vggt} & 1.1221 & 1.2198 & 34.9700 \\
$\pi$-Long & 0.0718 & \secondc{0.0847} & \secondc{2.5766}
& $\pi$-Long & \secondc{0.5287} & \secondc{0.7131} & 5.5125 \\
Scal3R~\cite{xie2026scal3r} & 0.1465 & 0.1428 & 4.0172
& Scal3R~\cite{xie2026scal3r} & 0.6521 & 0.8170 & \secondc{5.4277} \\
\textbf{Ours} & \firstc{\textbf{0.0591}} & \firstc{\textbf{0.0779}} & \firstc{\textbf{2.2940}}
& \textbf{Ours} & \firstc{\textbf{0.3852}} & \firstc{\textbf{0.5808}} & \firstc{\textbf{3.8230}} \\

\midrule
\multicolumn{4}{c|}{\textbf{TUM}: \texttt{freiburg2\_360\_kidnap}} 
& \multicolumn{4}{c}{\textbf{TUM}: \texttt{freiburg2\_desk}} \\
\cmidrule(lr){1-4} \cmidrule(lr){5-8}
\textbf{Method} & ATE $\downarrow$ & RTE $\downarrow$ & RRE $\downarrow$
& \textbf{Method} & ATE $\downarrow$ & RTE $\downarrow$ & RRE $\downarrow$ \\
\midrule
VGGT-Long~\cite{deng2025vggt} & \secondc{0.5252} & \secondc{0.9469} & 29.1949
& VGGT-Long~\cite{deng2025vggt} & 0.0946 & 0.3076 & 23.7105 \\
$\pi$-Long & 0.9643 & 1.2310 & \secondc{17.9083}
& $\pi$-Long & \secondc{0.0920} & \secondc{0.1263} & \secondc{2.1990} \\
Scal3R~\cite{xie2026scal3r} & 0.9654 & 1.1700 & 25.1519
& Scal3R~\cite{xie2026scal3r} & 0.1381 & 0.1389 & 3.3197 \\
\textbf{Ours} & \firstc{\textbf{0.2774}} & \firstc{\textbf{0.4017}} & \firstc{\textbf{6.9202}}
& \textbf{Ours} & \firstc{\textbf{0.0542}} & \firstc{\textbf{0.0850}} & \firstc{\textbf{1.9054}} \\

\midrule
\multicolumn{4}{c|}{\textbf{TUM}: \texttt{freiburg2\_desk\_with\_person}} 
& \multicolumn{4}{c}{\textbf{TUM}: \texttt{freiburg3\_sitting\_halfsphere}} \\
\cmidrule(lr){1-4} \cmidrule(lr){5-8}
\textbf{Method} & ATE $\downarrow$ & RTE $\downarrow$ & RRE $\downarrow$
& \textbf{Method} & ATE $\downarrow$ & RTE $\downarrow$ & RRE $\downarrow$ \\
\midrule
VGGT-Long~\cite{deng2025vggt} & 0.1696 & 0.4420 & 21.5829
& VGGT-Long~\cite{deng2025vggt} & 0.3068 & 0.3825 & 20.3508 \\
$\pi$-Long & \secondc{0.1158} & \secondc{0.1427} & \secondc{1.9198}
& $\pi$-Long & \secondc{0.1318} & \secondc{0.1840} & 13.1813 \\
Scal3R~\cite{xie2026scal3r} & 0.2148 & 0.2575 & 3.5409
& Scal3R~\cite{xie2026scal3r} & 0.2636 & 0.2879 & \firstc{\textbf{3.7372}} \\
\textbf{Ours} & \firstc{\textbf{0.1141}} & \firstc{\textbf{0.1382}} & \firstc{\textbf{1.9061}}
& \textbf{Ours} & \firstc{\textbf{0.1301}} & \firstc{\textbf{0.1790}} & \secondc{12.9796} \\

\midrule
\multicolumn{4}{c|}{\textbf{TUM}: \texttt{freiburg3\_sitting\_xyz}} 
& \multicolumn{4}{c}{\textbf{7Scenes}: \texttt{7scenes\_chess\_seq-03}} \\
\cmidrule(lr){1-4} \cmidrule(lr){5-8}
\textbf{Method} & ATE $\downarrow$ & RTE $\downarrow$ & RRE $\downarrow$
& \textbf{Method} & ATE $\downarrow$ & RTE $\downarrow$ & RRE $\downarrow$ \\
\midrule
VGGT-Long~\cite{deng2025vggt} & 0.1798 & 0.2670 & 35.8125
& VGGT-Long~\cite{deng2025vggt} & 0.0622 & 0.3158 & 22.5072 \\
$\pi$-Long & \secondc{0.1507} & 0.2150 & 2.4604
& $\pi$-Long & \secondc{0.0440} & \secondc{0.1356} & \secondc{17.2613} \\
Scal3R~\cite{xie2026scal3r} & 0.1640 & \secondc{0.2078} & \firstc{\textbf{2.2990}}
& Scal3R~\cite{xie2026scal3r} & 0.0717 & 0.2324 & 20.4186 \\
\textbf{Ours} & \firstc{\textbf{0.1346}} & \firstc{\textbf{0.1936}} & \secondc{2.4585}
& \textbf{Ours} & \firstc{\textbf{0.0418}} & \firstc{\textbf{0.1295}} & \firstc{\textbf{16.9348}} \\

\midrule
\multicolumn{4}{c|}{\textbf{7Scenes}: \texttt{7scenes\_fire\_seq-03}} 
& \multicolumn{4}{c}{\textbf{7Scenes}: \texttt{7scenes\_fire\_seq-04}} \\
\cmidrule(lr){1-4} \cmidrule(lr){5-8}
\textbf{Method} & ATE $\downarrow$ & RTE $\downarrow$ & RRE $\downarrow$
& \textbf{Method} & ATE $\downarrow$ & RTE $\downarrow$ & RRE $\downarrow$ \\
\midrule
VGGT-Long~\cite{deng2025vggt} & \firstc{\textbf{0.0480}} & \secondc{0.1401} & 26.0183
& VGGT-Long~\cite{deng2025vggt} & \firstc{\textbf{0.0502}} & 0.1412 & 21.7594 \\
$\pi$-Long & 0.0773 & 0.1574 & \secondc{25.9278}
& $\pi$-Long & 0.0577 & \secondc{0.1097} & \secondc{16.3985} \\
Scal3R~\cite{xie2026scal3r} & 0.0735 & 0.1831 & 27.5250
& Scal3R~\cite{xie2026scal3r} & 0.0567 & 0.1300 & 17.3516 \\
\textbf{Ours} & \secondc{0.0510} & \firstc{\textbf{0.1350}} & \firstc{\textbf{24.0130}}
& \textbf{Ours} & \secondc{0.0563} & \firstc{\textbf{0.1071}} & \firstc{\textbf{16.2720}} \\

\bottomrule
\end{tabular}
}
\end{table*}

\begin{table*}[t]
\caption{
\textbf{Representative camera pose estimation results, Part II.}
We report representative per-scene results for camera pose estimation, comparing LIST3R with strong subsequence-based methods, including VGGT-Long, $\pi$-Long, and Scal3R.
These sequences highlight cases where LIST3R achieves competitive or superior results across all three metrics.
Best and second-best results are highlighted.
}
\label{tab:selected_camera_pose_part2}
\centering
\renewcommand{\arraystretch}{1.02}
\setlength{\tabcolsep}{2.2pt}
\resizebox{\textwidth}{!}{
\begin{tabular}{@{}lccc|lccc@{}}
\toprule
\multicolumn{4}{c|}{\textbf{7Scenes}: \texttt{7scenes\_heads\_seq-01}} 
& \multicolumn{4}{c}{\textbf{7Scenes}: \texttt{7scenes\_office\_seq-02}} \\
\cmidrule(lr){1-4} \cmidrule(lr){5-8}
\textbf{Method} & {ATE (m) $\downarrow$} & {RTE (m) $\downarrow$} & {RRE (deg) $\downarrow$}
& \textbf{Method} & {ATE (m) $\downarrow$} & {RTE (m) $\downarrow$} & {RRE (deg) $\downarrow$} \\
\midrule
VGGT-Long~\cite{deng2025vggt} & 0.0444 & \secondc{0.1236} & 17.2139
& VGGT-Long~\cite{deng2025vggt} & 0.0931 & 0.3257 & 23.8709 \\
$\pi$-Long & \secondc{0.0380} & 0.1348 & \firstc{\textbf{8.5529}}
& $\pi$-Long & \secondc{0.0750} & \secondc{0.1355} & \firstc{\textbf{13.8847}} \\
Scal3R~\cite{xie2026scal3r} & 0.0593 & 0.1416 & 12.5768
& Scal3R~\cite{xie2026scal3r} & 0.1166 & 0.2095 & 20.3694 \\
\textbf{Ours} & \firstc{\textbf{0.0317}} & \firstc{\textbf{0.1140}} & \secondc{12.4969}
& \textbf{Ours} & \firstc{\textbf{0.0669}} & \firstc{\textbf{0.1205}} & \secondc{15.2410} \\

\midrule
\multicolumn{4}{c|}{\textbf{7Scenes}: \texttt{7scenes\_office\_seq-06}} 
& \multicolumn{4}{c}{\textbf{7Scenes}: \texttt{7scenes\_office\_seq-09}} \\
\cmidrule(lr){1-4} \cmidrule(lr){5-8}
\textbf{Method} & ATE $\downarrow$ & RTE $\downarrow$ & RRE $\downarrow$
& \textbf{Method} & ATE $\downarrow$ & RTE $\downarrow$ & RRE $\downarrow$ \\
\midrule
VGGT-Long~\cite{deng2025vggt} & 0.3759 & 0.4011 & 23.4726
& VGGT-Long~\cite{deng2025vggt} & 0.4368 & 0.5848 & 31.4176 \\
$\pi$-Long & \firstc{\textbf{0.0870}} & \secondc{0.1633} & \secondc{7.0507}
& $\pi$-Long & \secondc{0.1013} & \secondc{0.2111} & \secondc{9.7635} \\
Scal3R~\cite{xie2026scal3r} & 0.1146 & 0.2214 & 10.4250
& Scal3R~\cite{xie2026scal3r} & 0.1382 & 0.4333 & 14.6079 \\
\textbf{Ours} & \secondc{0.0926} & \firstc{\textbf{0.1616}} & \firstc{\textbf{7.0422}}
& \textbf{Ours} & \firstc{\textbf{0.0978}} & \firstc{\textbf{0.2077}} & \firstc{\textbf{9.6322}} \\

\midrule
\multicolumn{4}{c|}{\textbf{7Scenes}: \texttt{7scenes\_pumpkin\_seq-01}} 
& \multicolumn{4}{c}{\textbf{7Scenes}: \texttt{7scenes\_redkitchen\_seq-06}} \\
\cmidrule(lr){1-4} \cmidrule(lr){5-8}
\textbf{Method} & ATE $\downarrow$ & RTE $\downarrow$ & RRE $\downarrow$
& \textbf{Method} & ATE $\downarrow$ & RTE $\downarrow$ & RRE $\downarrow$ \\
\midrule
VGGT-Long~\cite{deng2025vggt} & 0.1649 & 0.2772 & 22.2682
& VGGT-Long~\cite{deng2025vggt} & 0.0922 & 0.2025 & 19.8909 \\
$\pi$-Long & \firstc{\textbf{0.1527}} & \secondc{0.2300} & \secondc{20.6697}
& $\pi$-Long & \secondc{0.0891} & \secondc{0.1430} & \secondc{14.3598} \\
Scal3R~\cite{xie2026scal3r} & 0.2281 & 0.2978 & 24.4133
& Scal3R~\cite{xie2026scal3r} & 0.1039 & 0.2123 & 20.8689 \\
\textbf{Ours} & \secondc{0.1582} & \firstc{\textbf{0.2295}} & \firstc{\textbf{20.3426}}
& \textbf{Ours} & \firstc{\textbf{0.0671}} & \firstc{\textbf{0.1233}} & \firstc{\textbf{13.8664}} \\

\midrule
\multicolumn{4}{c|}{\textbf{7Scenes}: \texttt{7scenes\_redkitchen\_seq-12}} 
& \multicolumn{4}{c}{\textbf{7Scenes}: \texttt{7scenes\_stairs\_seq-04}} \\
\cmidrule(lr){1-4} \cmidrule(lr){5-8}
\textbf{Method} & ATE $\downarrow$ & RTE $\downarrow$ & RRE $\downarrow$
& \textbf{Method} & ATE $\downarrow$ & RTE $\downarrow$ & RRE $\downarrow$ \\
\midrule
VGGT-Long~\cite{deng2025vggt} & \secondc{0.0924} & 0.2219 & 11.0076
& VGGT-Long~\cite{deng2025vggt} & 0.0643 & 0.1238 & 29.8025 \\
$\pi$-Long & 0.1524 & \secondc{0.1860} & \secondc{6.1936}
& $\pi$-Long & \secondc{0.0609} & \secondc{0.0745} & 24.4008 \\
Scal3R~\cite{xie2026scal3r} & 0.1118 & 0.2368 & 8.4014
& Scal3R~\cite{xie2026scal3r} & 0.0962 & 0.1275 & \firstc{\textbf{22.0758}} \\
\textbf{Ours} & \firstc{\textbf{0.0867}} & \firstc{\textbf{0.1691}} & \firstc{\textbf{5.7333}}
& \textbf{Ours} & \firstc{\textbf{0.0552}} & \firstc{\textbf{0.0742}} & \secondc{24.3354} \\

\midrule
\multicolumn{4}{c|}{\textbf{ETH3D}: \texttt{ceiling\_1}} 
& \multicolumn{4}{c}{\textbf{ETH3D}: \texttt{desk\_changing\_1}} \\
\cmidrule(lr){1-4} \cmidrule(lr){5-8}
\textbf{Method} & ATE $\downarrow$ & RTE $\downarrow$ & RRE $\downarrow$
& \textbf{Method} & ATE $\downarrow$ & RTE $\downarrow$ & RRE $\downarrow$ \\
\midrule
VGGT-Long~\cite{deng2025vggt} & 1.6208 & 2.1122 & 43.9541
& VGGT-Long~\cite{deng2025vggt} & 0.8053 & 0.8983 & 30.3645 \\
$\pi$-Long & \secondc{0.7010} & \secondc{0.5572} & \firstc{\textbf{10.3401}}
& $\pi$-Long & \secondc{0.1569} & \secondc{0.1489} & \secondc{5.0824} \\
Scal3R~\cite{xie2026scal3r} & 1.3272 & 1.3069 & 14.9191
& Scal3R~\cite{xie2026scal3r} & 0.2897 & 0.2456 & 10.8936 \\
\textbf{Ours} & \firstc{\textbf{0.3701}} & \firstc{\textbf{0.4812}} & \secondc{12.7321}
& \textbf{Ours} & \firstc{\textbf{0.0712}} & \firstc{\textbf{0.1109}} & \firstc{\textbf{4.0349}} \\

\midrule
\multicolumn{4}{c|}{\textbf{ETH3D}: \texttt{kidnap\_1}} 
& \multicolumn{4}{c}{\textbf{ETH3D}: \texttt{motion\_1}} \\
\cmidrule(lr){1-4} \cmidrule(lr){5-8}
\textbf{Method} & ATE $\downarrow$ & RTE $\downarrow$ & RRE $\downarrow$
& \textbf{Method} & ATE $\downarrow$ & RTE $\downarrow$ & RRE $\downarrow$ \\
\midrule
VGGT-Long~\cite{deng2025vggt} & \secondc{0.5270} & 1.0618 & 41.4875
& VGGT-Long~\cite{deng2025vggt} & 0.1463 & 0.2875 & 11.8862 \\
$\pi$-Long & 0.6260 & 0.8707 & 37.5038
& $\pi$-Long & \secondc{0.1286} & \secondc{0.1259} & \secondc{2.0350} \\
Scal3R~\cite{xie2026scal3r} & 0.5993 & \secondc{0.8663} & \secondc{33.8688}
& Scal3R~\cite{xie2026scal3r} & 0.2119 & 0.2067 & 2.5391 \\
\textbf{Ours} & \firstc{\textbf{0.0535}} & \firstc{\textbf{0.2291}} & \firstc{\textbf{26.9660}}
& \textbf{Ours} & \firstc{\textbf{0.0953}} & \firstc{\textbf{0.1065}} & \firstc{\textbf{1.8449}} \\

\bottomrule
\end{tabular}
}
\end{table*}

\begin{table}[t]
\caption{
\textbf{Selected per-scene point cloud reconstruction results.}
We report scenes where LIST3R achieves the best rank on all five metrics, or obtains at least three best ranks while all five metrics remain within the top two.
CUT3R and TTT3R are excluded from this per-scene comparison.
Best and second-best results are highlighted.
}
\label{tab:selected_point_cloud_quality}
\centering
\renewcommand{\arraystretch}{1.02}
\setlength{\tabcolsep}{3.5pt}
\resizebox{\linewidth}{!}{
\begin{tabular}{@{}lccccc@{}}
\toprule
\multicolumn{6}{c}{\textbf{ETH3D}: \texttt{ceiling\_1}} \\
\cmidrule(lr){1-6}
\textbf{Method}
& {Acc. (cm) $\downarrow$}
& {Comp. (cm) $\downarrow$}
& {Chamfer (cm) $\downarrow$}
& {Normal Consistency $\uparrow$}
& {F@5cm $\uparrow$} \\
\midrule
VGGT-Long~\cite{deng2025vggt}
& 33.063 & 30.862 & 31.962 & 0.588 & 12.164 \\
$\pi$-Long
& \secondc{18.675} & \secondc{6.430} & \secondc{12.552} & \firstc{\textbf{0.668}} & \firstc{\textbf{27.980}} \\
Scal3R~\cite{xie2026scal3r}
& 44.170 & 26.477 & 35.323 & 0.577 & 10.400 \\
\textbf{Ours}
& \firstc{\textbf{14.474}} & \firstc{\textbf{5.530}} & \firstc{\textbf{10.002}} & \secondc{0.665} & \secondc{27.747} \\

\midrule
\multicolumn{6}{c}{\textbf{ETH3D}: \texttt{desk\_changing\_1}} \\
\cmidrule(lr){1-6}
\textbf{Method}
& {Acc. (cm) $\downarrow$}
& {Comp. (cm) $\downarrow$}
& {Chamfer (cm) $\downarrow$}
& {Normal Consistency $\uparrow$}
& {F@5cm $\uparrow$} \\
\midrule
VGGT-Long~\cite{deng2025vggt}
& 21.018 & 6.601 & 13.810 & 0.624 & 24.384 \\
$\pi$-Long
& \secondc{4.852} & \firstc{\textbf{2.768}} & \secondc{3.810} & \secondc{0.773} & \secondc{74.274} \\
Scal3R~\cite{xie2026scal3r}
& 10.603 & 20.191 & 15.397 & 0.652 & 37.284 \\
\textbf{Ours}
& \firstc{\textbf{3.421}} & \secondc{2.801} & \firstc{\textbf{3.111}} & \firstc{\textbf{0.798}} & \firstc{\textbf{83.756}} \\

\midrule
\multicolumn{6}{c}{\textbf{ETH3D}: \texttt{kidnap\_1}} \\
\cmidrule(lr){1-6}
\textbf{Method}
& {Acc. (cm) $\downarrow$}
& {Comp. (cm) $\downarrow$}
& {Chamfer (cm) $\downarrow$}
& {Normal Consistency $\uparrow$}
& {F@5cm $\uparrow$} \\
\midrule
VGGT-Long~\cite{deng2025vggt}
& \secondc{41.756} & \secondc{10.545} & \secondc{26.151} & 0.542 & 13.594 \\
$\pi$-Long
& 51.807 & 13.590 & 32.698 & \secondc{0.662} & \secondc{17.518} \\
Scal3R~\cite{xie2026scal3r}
& 53.220 & 21.790 & 37.505 & 0.587 & 11.163 \\
\textbf{Ours}
& \firstc{\textbf{13.846}} & \firstc{\textbf{4.706}} & \firstc{\textbf{9.276}} & \firstc{\textbf{0.763}} & \firstc{\textbf{47.580}} \\

\midrule
\multicolumn{6}{c}{\textbf{ETH3D}: \texttt{plant\_scene\_3}} \\
\cmidrule(lr){1-6}
\textbf{Method}
& {Acc. (cm) $\downarrow$}
& {Comp. (cm) $\downarrow$}
& {Chamfer (cm) $\downarrow$}
& {Normal Consistency $\uparrow$}
& {F@5cm $\uparrow$} \\
\midrule
VGGT-Long~\cite{deng2025vggt}
& 13.198 & 8.167 & 10.682 & 0.692 & 38.917 \\
$\pi$-Long
& 8.620 & \secondc{3.827} & \secondc{6.224} & \secondc{0.722} & \secondc{73.460} \\
Scal3R~\cite{xie2026scal3r}
& \firstc{\textbf{7.748}} & 11.606 & 9.677 & 0.685 & 44.982 \\
\textbf{Ours}
& \secondc{8.218} & \firstc{\textbf{3.209}} & \firstc{\textbf{5.713}} & \firstc{\textbf{0.726}} & \firstc{\textbf{76.330}} \\

\midrule
\multicolumn{6}{c}{\textbf{NRGBD}: \texttt{grey\_white\_room}} \\
\cmidrule(lr){1-6}
\textbf{Method}
& {Acc. (cm) $\downarrow$}
& {Comp. (cm) $\downarrow$}
& {Chamfer (cm) $\downarrow$}
& {Normal Consistency $\uparrow$}
& {F@5cm $\uparrow$} \\
\midrule
VGGT-Long~\cite{deng2025vggt}
& 7.963 & 4.155 & 6.059 & 0.736 & 52.381 \\
$\pi$-Long
& \secondc{4.891} & \secondc{3.124} & \secondc{4.007} & \secondc{0.836} & \secondc{70.050} \\
Scal3R~\cite{xie2026scal3r}
& 6.325 & 3.163 & 4.744 & 0.784 & 66.517 \\
\textbf{Ours}
& \firstc{\textbf{4.023}} & \firstc{\textbf{2.694}} & \firstc{\textbf{3.358}} & \firstc{\textbf{0.870}} & \firstc{\textbf{84.136}} \\

\midrule
\multicolumn{6}{c}{\textbf{NRGBD}: \texttt{kitchen}} \\
\cmidrule(lr){1-6}
\textbf{Method}
& {Acc. (cm) $\downarrow$}
& {Comp. (cm) $\downarrow$}
& {Chamfer (cm) $\downarrow$}
& {Normal Consistency $\uparrow$}
& {F@5cm $\uparrow$} \\
\midrule
VGGT-Long~\cite{deng2025vggt}
& 16.605 & 8.309 & 12.457 & 0.792 & 34.788 \\
$\pi$-Long
& \secondc{5.947} & \secondc{4.709} & \secondc{5.328} & \firstc{\textbf{0.895}} & \secondc{55.989} \\
Scal3R~\cite{xie2026scal3r}
& 7.789 & 5.818 & 6.803 & 0.843 & 52.666 \\
\textbf{Ours}
& \firstc{\textbf{4.825}} & \firstc{\textbf{3.711}} & \firstc{\textbf{4.268}} & \secondc{0.887} & \firstc{\textbf{69.084}} \\

\bottomrule
\end{tabular}
}
\end{table}

\subsection{Additional Quantitative Results}
\label{app:quantitative_results}

We provide additional representative per-scene quantitative results for camera trajectory estimation and point cloud reconstruction.
These results complement the aggregate comparisons reported in the main paper and provide a more detailed view of LIST3R on individual long sequences.

Tables~\ref{tab:selected_camera_pose_part1} and~\ref{tab:selected_camera_pose_part2} report representative camera pose estimation results.
We compare LIST3R with strong subsequence-based baselines, including VGGT-Long, $\pi$-Long, and Scal3R.
The selected sequences cover different datasets and motion patterns, including indoor trajectories, large camera rotations, and challenging revisit cases.
Across these examples, LIST3R achieves competitive or superior performance on ATE, RTE, and RRE, showing its ability to maintain reliable camera trajectory consistency after subsequence association.

Table~\ref{tab:selected_point_cloud_quality} reports representative point cloud reconstruction results on selected long sequences.
These examples cover both ETH3D and NRGBD scenes, including indoor environments with large planar structures, object-rich regions, and challenging camera motion.
Compared with the subsequence-based baselines, LIST3R produces more reliable fused geometry on these representative scenes.
The improvements are reflected across complementary reconstruction metrics, including accuracy, completion, Chamfer distance, normal consistency, and F@5cm.
This suggests that the instance-aware association and merging strategy not only improves camera alignment, but also helps preserve more accurate and complete scene geometry after cross-subsequence integration.

\subsection{Additional Camera Trajectory Visualizations}
\label{app:trajectory_visualizations}

\begin{figure*}[t]
  \centering
  \includegraphics[width=0.95\textwidth]{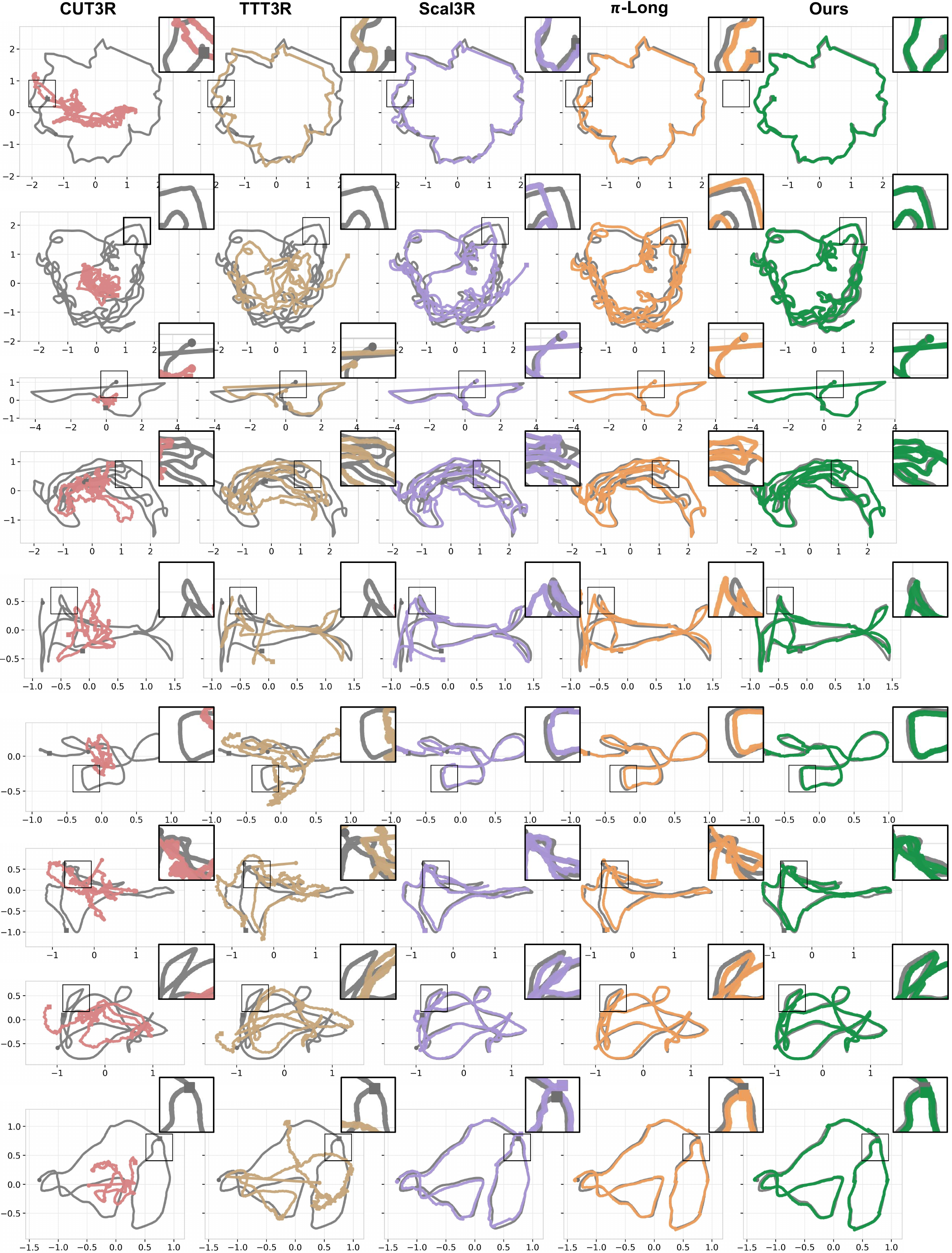}
  \vspace{-0.6em}
  \caption{
  \textbf{Additional visual comparisons of camera trajectory estimation results.}
  We present per-scene trajectory comparisons between the estimated camera trajectories and the ground truth.
  Gray curves denote the ground-truth trajectories, while colored curves denote the predictions of different methods.
  The zoomed regions highlight challenging trajectory segments where LIST3R better preserves global consistency and reduces drift.
  }
  \label{fig:supp_camera_pose}
  \vspace{-1.0em}
\end{figure*}

We provide additional camera trajectory visualizations to complement the quantitative pose results.
Fig.~\ref{fig:supp_camera_pose} compares the estimated trajectories of different methods with the ground-truth trajectories on representative long sequences.
The zoomed regions highlight local trajectory differences around challenging segments, such as revisits, turns, and accumulated drift.
Compared with the baselines, LIST3R more consistently follows the global trajectory structure, indicating that instance-aware long-range associations provide useful constraints for reducing drift.

\subsection{Additional Point Cloud Reconstruction Visualizations}
\label{app:reconstruction_visualizations}

\begin{figure*}[t]
  \centering
  \includegraphics[width=0.95\textwidth]{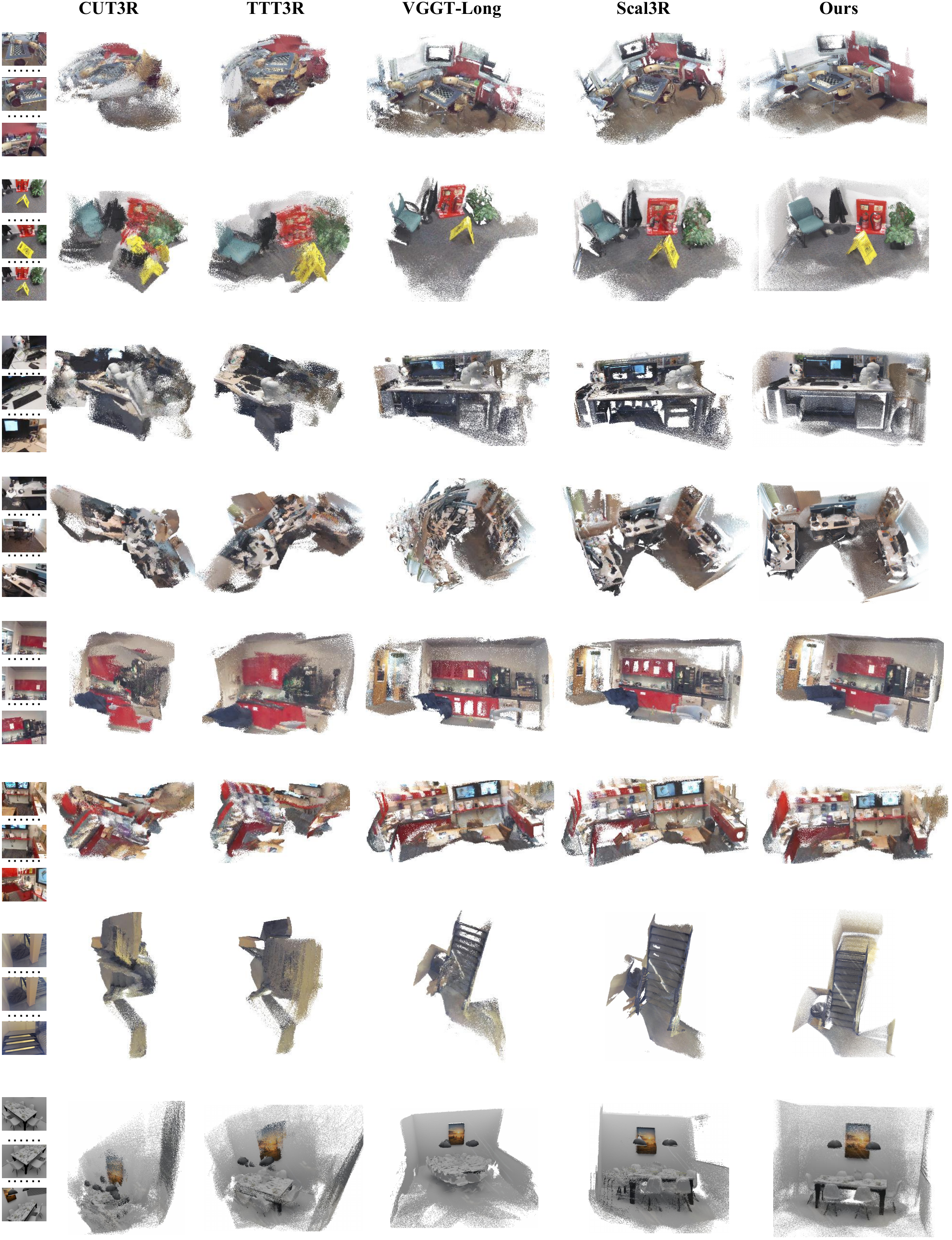}
  \vspace{-0.6em}
  \caption{
\textbf{Additional visual comparisons of point cloud reconstruction results, Part I.}
We present more per-scene visual comparisons of point cloud reconstruction.
These examples highlight cases where LIST3R recovers more complete geometry and preserves clearer scene structures.
}
  \label{fig:supp_cloud_vis1}
  \vspace{-1.0em}
\end{figure*}

\begin{figure*}[t]
  \centering
  \includegraphics[width=0.95\textwidth]{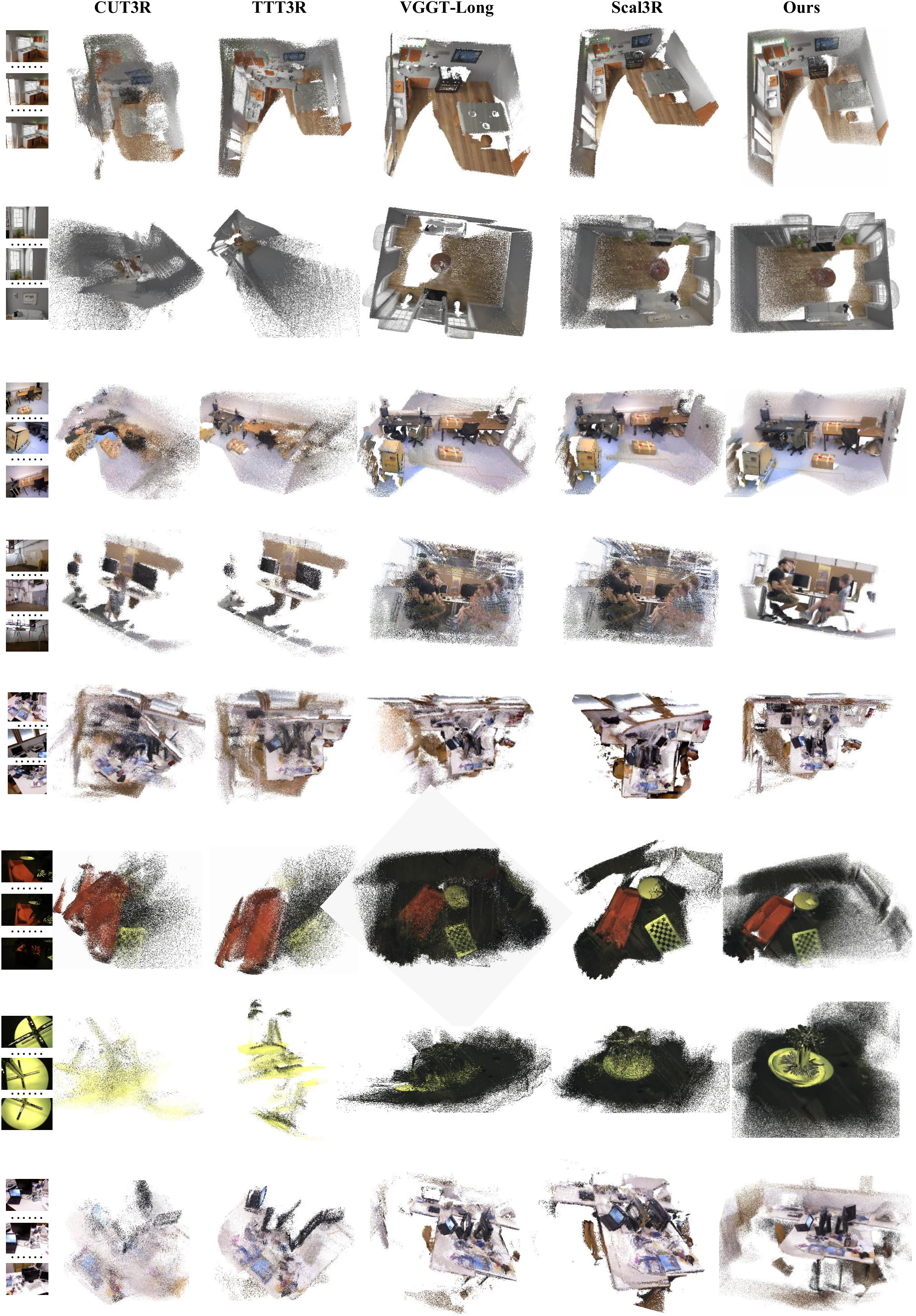}
  \vspace{-0.6em}
  \caption{
\textbf{Additional visual comparisons of point cloud reconstruction results, Part II.}
We present more per-scene visual comparisons of point cloud reconstruction.
These examples highlight cases where LIST3R recovers more complete geometry and preserves clearer scene structures.
}
  \label{fig:supp_cloud_vis2}
  \vspace{-1.0em}
\end{figure*}

\begin{figure*}[t]
  \centering
  \includegraphics[width=0.95\textwidth]{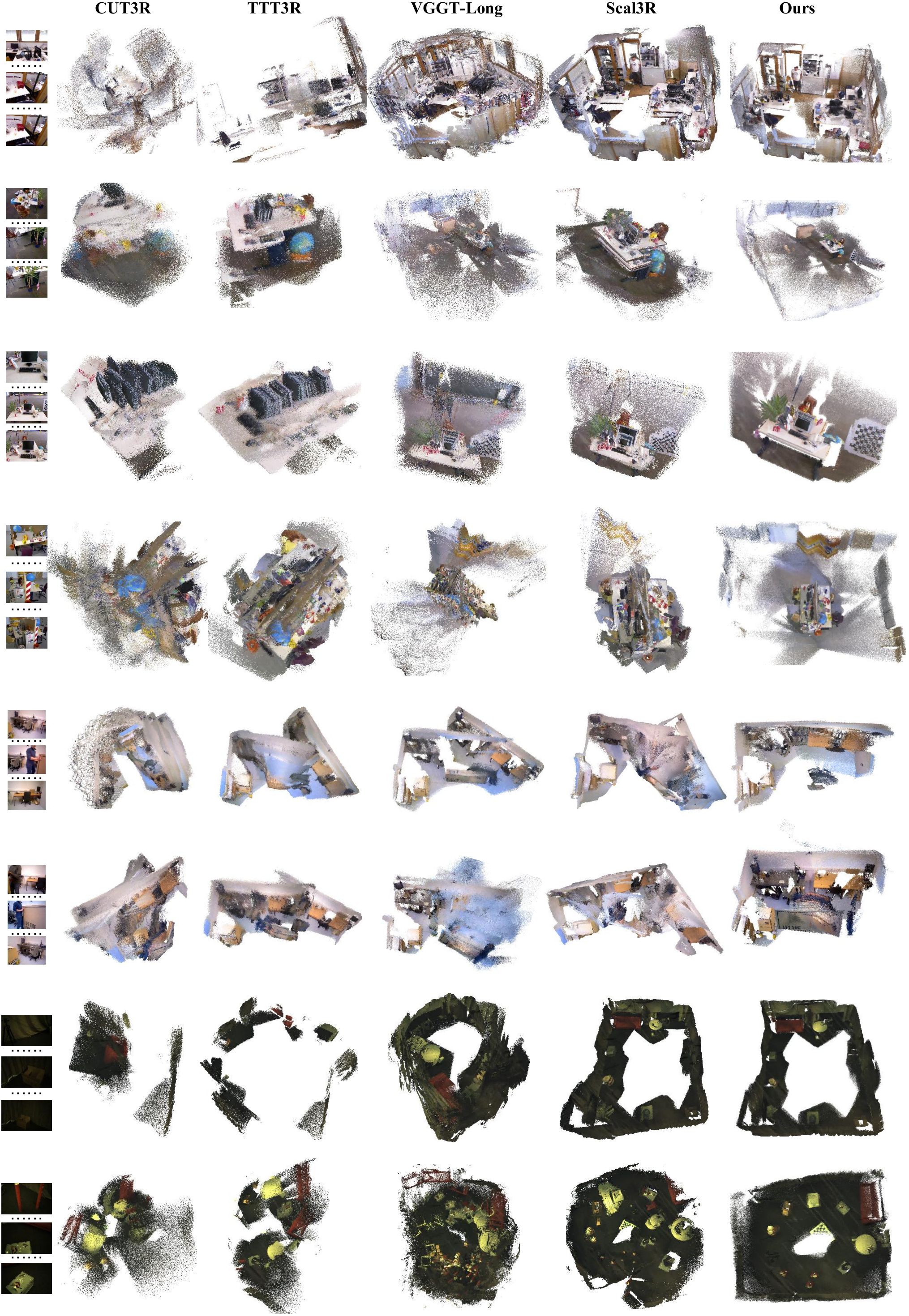}
  \vspace{-0.6em}
  \caption{
\textbf{Additional visual comparisons of point cloud reconstruction results, Part III.}
We present more per-scene visual comparisons of point cloud reconstruction.
These examples highlight cases where LIST3R recovers more complete geometry and preserves clearer scene structures.
}
  \label{fig:supp_cloud_vis3}
  \vspace{-1.0em}
\end{figure*}

We further provide additional point cloud reconstruction visualizations in Figs.~\ref{fig:supp_cloud_vis1}, \ref{fig:supp_cloud_vis2}, and \ref{fig:supp_cloud_vis3}.
These examples cover representative long sequences with different scene layouts, object distributions, and camera motions.
The visual comparisons show that LIST3R produces more coherent fused geometry after cross-subsequence integration.
In particular, the reconstructed point clouds tend to preserve clearer object boundaries and more complete scene structures, while reducing visible misalignment artifacts caused by unreliable subsequence associations.

\section{Limitations}
\label{app:limitations}

LIST3R has several limitations. First, its performance depends on the quality of instance discovery and mask tracking. Since the instance library provides anchors for revisit discovery and subsequence association, unreliable instance observations may affect the robustness of the subsequent reconstruction process.
Second, our method assumes that the scene contains stable and repeatedly observable object instances. This assumption is suitable for many indoor and object-rich environments, but may be less effective in object-sparse scenes, repetitive corridors, or regions dominated by large weakly structured surfaces, where persistent and discriminative anchors are limited.
In principle, the proposed instance-anchor formulation can be extended to broader reconstruction scenarios. However, large-scale outdoor scenes involve more diverse viewpoints, illumination changes, object scales, and dynamic factors. Addressing these challenges may require additional strategies for robust instance discovery, geometric verification, and adaptive anchor selection, which we leave as future work.
Finally, long-sequence 3D reconstruction may raise privacy concerns when applied to private or sensitive environments. Practical use should respect data ownership, consent, and dataset licenses.

\section{Broader Impacts}
\label{app:broader_impacts}

This work aims to improve long-sequence 3D reconstruction by using persistent object instances as anchors for organizing visual observations across time. The potential positive impacts include more scalable and consistent 3D reconstruction for robotics, augmented and virtual reality, digital mapping, cultural heritage preservation, and embodied scene understanding. By improving reconstruction consistency over long videos, the method may help systems better understand large-scale environments and support downstream applications that require reliable geometry and scene structure.

At the same time, improved 3D reconstruction may raise privacy and security concerns when applied to private, sensitive, or restricted environments. For example, reconstructing indoor spaces from long videos could reveal personal belongings, spatial layouts, or other sensitive information. Such technology could also be misused for unintended surveillance or unauthorized mapping. To mitigate these risks, the method should be used with appropriate consent, respect for dataset licenses and terms of use, and care when releasing reconstructed scenes. In this work, experiments are conducted on existing public benchmarks, and we do not release sensitive private-scene data.

\clearpage


\end{document}